\documentclass[
  a4paper,
  fontsize=10pt,
]{kaohandt}

\usepackage{calc}
\usepackage[english]{babel}
\usepackage[utf8]{inputenc}
\usepackage[T1]{fontenc}
\usepackage{amsmath,amssymb}
\usepackage{graphicx}
\usepackage{subcaption}
\usepackage{placeins}
\usepackage{booktabs}
\usepackage{xcolor}
\usepackage{listings}
\usepackage{enumitem}
\usepackage{tikz}
\usepackage[ruled,linesnumbered]{algorithm2e}
\SetAlgoNlRelativeSize{0}
\SetAlgoSkip{medskip}

\usepackage[maxbibnames=6,minbibnames=6]{kaobiblio}
\usepackage{kaotheorems}
\usepackage{kaorefs}

\definecolor{theaLinkBlue}{HTML}{4A6480}
\hypersetup{citecolor=theaLinkBlue, urlcolor=theaLinkBlue, filecolor=theaLinkBlue}

\makeatletter
\renewcommand\marginskip[1]{%
  \marginpar{\@margin@par\setlength{\@tempskipa}{#1-\baselineskip}\vspace{\@tempskipa}}%
}
\makeatother

\definecolor{gapaccent}{HTML}{55524E}%
\definecolor{gapteal}{HTML}{6B8583}
\definecolor{gapmauve}{HTML}{8D7B80}%
\definecolor{gapinkbg}{HTML}{F9F9F8}%
\definecolor{gapbg}{HTML}{F7F7F7}%

\newtcolorbox{kaoboxA}[3][]{
  enhanced, breakable,
  before skip=1.1\topskip, after skip=1.1\topskip,
  left=12pt, right=12pt, top=5pt, bottom=6pt,
  lefttitle=12pt, toptitle=4pt, bottomtitle=2pt,
  boxrule=0pt, frame hidden, sharp corners,
  colback=#3, colbacktitle=#3,
  borderline west={2.5pt}{0pt}{#2},
  coltitle=#2, fonttitle=\large\bfseries, titlerule=0pt,
  #1
}

\newtcolorbox{kaoboxB}[1][]{
  enhanced, breakable,
  before skip=1.4\topskip, after skip=1.4\topskip,
  left=12pt, right=12pt, top=7pt, bottom=8pt,
  lefttitle=12pt, toptitle=6pt, bottomtitle=4pt,
  rounded corners, arc=2.5pt, boxrule=0.6pt,
  colback=gapbg, colbacktitle=gapbg, colframe=black!18,
  coltitle=black, fonttitle=\Large\bfseries, titlerule=0pt,
  #1
}

\usepackage{xspace}

\renewcommand{\eg}{e.g.\@\xspace}
\newcommand{\thea}{\textsc{Thea}\xspace}
\definecolor{wentaocolor}{HTML}{7B68EE}

\DeclareRobustCommand{\code}[1]{%
  {\setlength{\fboxsep}{1.5pt}%
   \raisebox{0pt}[0pt][0pt]{\colorbox{codebg}{\vphantom{gk}\ttfamily#1}}}}

\definecolor{codebg}{HTML}{F8F8F7}       %
\definecolor{codekw}{HTML}{2B679D}       %
\definecolor{codecomment}{HTML}{A39E93}  %
\definecolor{codestring}{HTML}{2E8B6F}   %
\definecolor{codebuiltin}{HTML}{BC6448}  %
\definecolor{codenum}{HTML}{C2BEB6}      %

\definecolor{contextresidentbg}{HTML}{EFECE6}
\definecolor{contextresidentink}{HTML}{6E675C}
\definecolor{contextrefreshedbg}{HTML}{EBF0EE}
\definecolor{contextrefreshedink}{HTML}{4E6361}
\definecolor{contextaccumulatedbg}{HTML}{EAEAE8}
\definecolor{contextaccumulatedink}{HTML}{55524E}

\lstdefinelanguage{json}{
  basicstyle=\ttfamily\small,
  showstringspaces=false,
  breaklines=true,
  morestring=[b]",
  morecomment=[l]{//},
  keywords={true,false,null},
  keywordstyle=\color{codebuiltin}\bfseries,
}

\lstdefinestyle{thea}{
  language=Python,
  basicstyle=\ttfamily\small,
  keywordstyle=\color{codekw}\bfseries,
  commentstyle=\color{codecomment}\itshape,
  stringstyle=\color{codestring},
  emphstyle=\color{codebuiltin},
  emph={True,False,None,self},
  breaklines=true,
  backgroundcolor=\color{codebg},
  frame=single,
  framerule=0pt,
  rulecolor=\color{codebg},
  framesep=8pt,
  framextopmargin=6pt,
  framexbottommargin=6pt,
  xleftmargin=2.2em,
  framexleftmargin=2.2em,
  aboveskip=1.2em,
  belowskip=1.2em,
  numbers=left,
  numberstyle=\ttfamily\scriptsize\color{codenum},
  numbersep=10pt,
  tabsize=4,
  showstringspaces=false,
  captionpos=b,
}
\AtBeginDocument{%
  \captionsetup[lstlisting]{
    format=plain,
    font=small,
    labelfont=bf,
    labelsep=period,
    justification=raggedright,
    singlelinecheck=no,
    aboveskip=4pt,
    belowskip=0pt,
  }%
  \captionsetup[table]{labelsep=period}%
  \captionsetup[longtable]{labelsep=period}%
}

\lstdefinelanguage{markdown}{
  sensitive=true,
  morekeywords={name,description},
  morecomment=[l][\color{codecomment}]{---},
  moredelim=[l][\bfseries\color{codebuiltin}]{\#\#},
}

\title{\textbf{\mbox{Towards the Harness of Embodied Agents}}}
\author{%
  \large
  Qi Wang$^{\dagger}$, Tianyi Wang$^{\dagger}$, Chengyang Li$^{\dagger}$, Shikun Ban, Yurun Chen,\\[4pt]
  \large Yizhong Ge, Jason Qin, Chengtai Li$^{\dagger}$, Wentao Zhu$^{\dagger}$\\[8pt]
  \large Eastern Institute of Technology, Ningbo\\[6pt]
  \normalsize $^{\dagger}$\,Core contributors%
}
\date{\large Technical Report, July 2026}

\RedeclareSectionCommand[beforeskip=-6ex plus -1ex minus -.2ex,
                         afterskip=3.4ex plus .2ex]{section}
\RedeclareSectionCommand[beforeskip=-5ex plus -1ex minus -.2ex,
                         afterskip=2.4ex plus .2ex]{subsection}
\RedeclareSectionCommand[beforeskip=-4ex plus -1ex minus -.2ex,
                         afterskip=1.9ex plus .2ex]{subsubsection}

\newpagestyle{outerpagenum.scrheadings}{%
	{}%
	{}%
	{}%
}{%
	{\makebox[0pt][l]{\makebox[\headtotal][r]{\thepage}}}%
	{\makebox[0pt][l]{\makebox[\headtotal][r]{\thepage}}}%
	{\makebox[0pt][l]{\makebox[\headtotal][r]{\thepage}}}%
}
\begin{document}
\linespread{1.09}\selectfont

\maketitle

\marginnote[62pt]{%
  \textbf{Project page}\\[2pt]
  \url{https://eit-hai.github.io/thea}\\[7pt]
  \textbf{Code}\\[2pt]
  \url{https://github.com/EIT-HAI/Thea}%
}
\begin{kaoboxB}[title=Abstract, frame hidden, boxrule=0pt, before skip=4\topskip, arc=4pt, toptitle=11pt, bottom=13pt]
The success of coding agents has established the \emph{harness} as a paradigm: what an agent achieves depends not on the model alone, but on the infrastructure around it.
We ask whether the same paradigm extends to embodied agents in the physical world.
We present \thea, a harness in which an agentic loop orchestrates robot capabilities, each wrapped as a callable tool.
It inherits the core components of coding agents, modified as the physical world requires.
The world, however, withholds two abilities that software grants for free: reading the state of the world, and judging the outcome of an action.
To bridge these gaps, \thea introduces \emph{Scene Graph as Context}, a persistent, symbolic representation of the world, and \emph{Evaluation as Exit Codes}, which detects when an action should terminate, judges whether it succeeded, and on failure diagnoses the cause.
Together they close the loop between the agent and the physical world.
Rich behaviors then emerge from the composition of tools, and the closed loop carries long-horizon tasks to completion in real environments.

\end{kaoboxB}

\setcounter{tocdepth}{2}%
\setkomafont{sectionentry}{\normalfont\bfseries}%
\tableofcontents
\clearpage%

\section{Introduction}
\label{sec:intro}

\begin{figure*}[t!]
  \centering
  \vspace*{-8pt}
  \includegraphics[width=\dimexpr\textwidth+\marginparsep+\marginparwidth\relax]{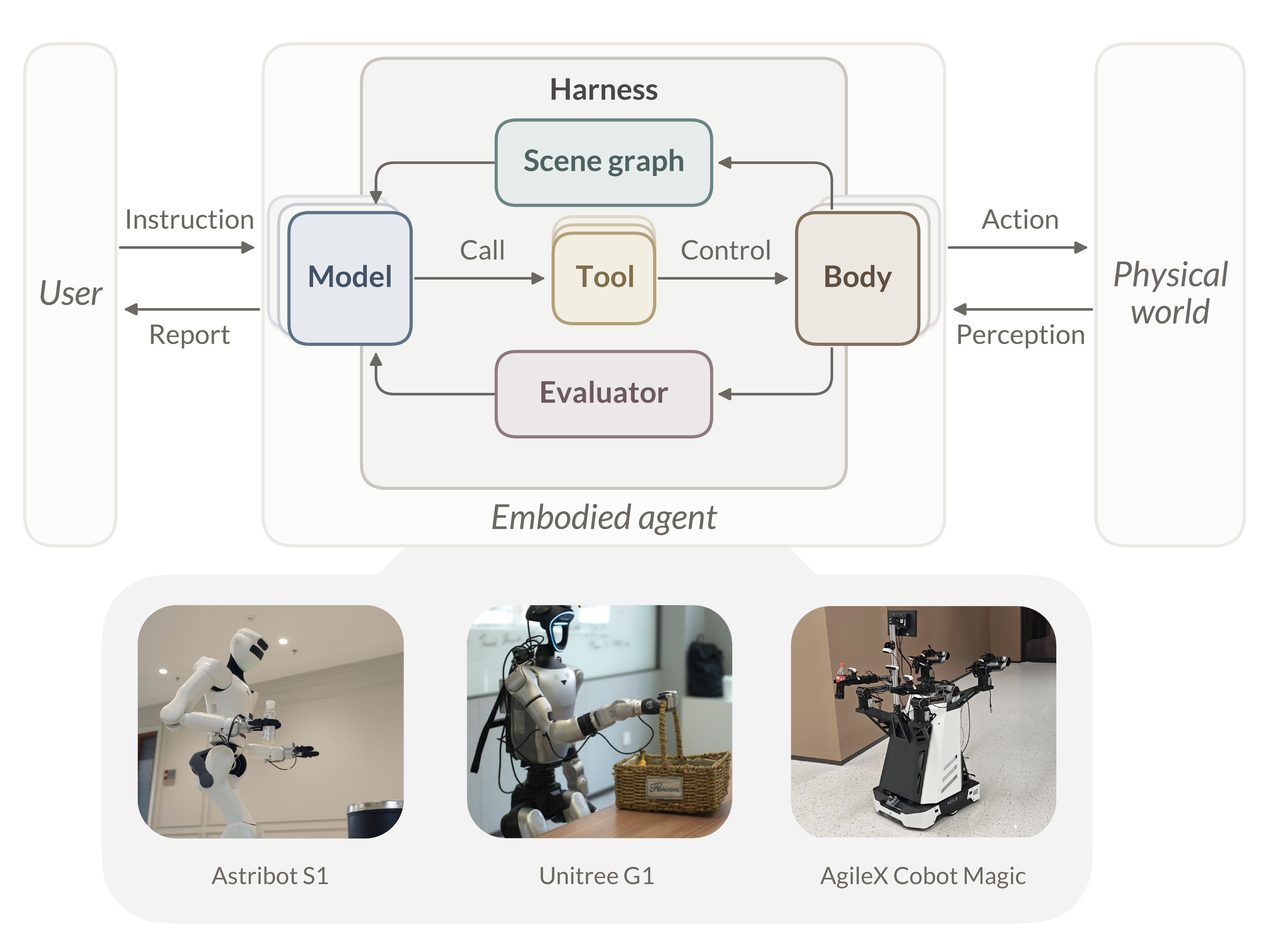}
  \refstepcounter{figure}
  \label{fig:overview}
  \vspace{2pt}
  \parbox{\dimexpr\textwidth+\marginparsep+\marginparwidth\relax}{\raggedright\small\textbf{Figure~\thefigure.} Overview of \thea. The harness wires a model and a body, each swappable, into an embodied agent. The model calls tools to control the body, and reads back the resulting world state from the scene graph and the outcome from the evaluator, closing the loop. The harness ports across embodiments, and we validate \thea on three different robot platforms.\par}
\end{figure*}

The rise of \emph{coding agents}~\sidecite{anthropic2026claude,openai2026codex,jimenez2024swebench} has reshaped software engineering.
Rather than produce correct code in a single pass, they write, test, read the error, fix, and test again, converging on correctness through a closed loop.
The insight is not a better model, but a better \emph{harness}~\sidecite{hashimoto2026harness,lopopolo2026harness}: the surrounding infrastructure that grounds the model in its environment, mediates its actions through tools, and closes the loop between intent and outcome through evaluation and recovery.
Complex behaviors emerge from the interaction between a simple agentic loop~\sidecite{anthropic2026agentloop} and its tools.
Plug in new tools and the same architecture that writes code also produces documents, queries external services, and conducts open-ended tasks with the user; capabilities arrive without changing the model or the loop.
The harness has become a paradigm~\sidecite{weng2026harness}.

All of this, however, unfolds in digital environments.
In the physical world, the building blocks are in place. Embodied foundation models now perform a range of tasks with growing competence, whether navigating cluttered rooms or manipulating everyday objects~\sidecite{brohan2023rt1,octo2024,black2025pi05}.
Yet a collection of strong atomic capabilities does not constitute a useful robot.
What remains open is \emph{orchestration}: weaving these capabilities into long-horizon behavior in a dynamic world shared with the people it serves.
This is exactly the problem coding agents have proved tractable in the digital world.
Their answer is not to wait for an omnipotent model, but to build a carefully engineered harness that amplifies what the model can achieve.
Does the same paradigm, then, hold in the physical world?

An agent is, in essence, a closed loop of perception and action: it perceives the world, acts on it, and perceives the result.
The architecture of coding agents provides a natural blueprint.
At each turn the language model selects a tool (\eg read a file, run a test, edit code), observes the result, and decides the next action.
Each robot capability (\eg navigation, manipulation) can likewise become a callable tool; a language model orchestrates them through the same agentic loop.
The design decouples orchestration from execution. The agentic loop accesses policies and platforms only through their tool interfaces, remaining agnostic to their internals.
Complex, long-horizon tasks need not be solved monolithically; the agentic loop addresses them step by step, and new behaviors emerge from flexible composition of tools.

At first glance, transferring this blueprint to the physical world should be a simple matter of redefining the tools.
It is not.
The physical world is far more complex than software.
Closing the loop demands two abilities: reading the state of the world, and judging the outcome of an action.
Software grants both for free; the physical world grants neither.

\begin{kaoboxA}[title=Gap 1: Readability of the World]{gapteal}{gapinkbg}
A software environment is a \emph{designed artifact}, built by humans for machines to read and execute.
Source code is text; a language model can read it, grep it, diff it, reason about it.
A coding agent perceives it simply by reading, because the environment \emph{is already structured, symbolic, and persistent}.
At any moment, the agent can take in the entire state, exact and complete.

The physical world is not a designed artifact. It simply exists.
An embodied agent perceives by sensing. Its input is a 30\,fps stream of RGB-D frames, continuous, high-dimensional, and unstructured.
The agent sees only a local, momentary slice of the world, never the whole.
It must piece the whole together frame by frame and hold it in memory; the world keeps no record to consult.
A coding agent's ``codebase'' is free; a physical agent's must be constructed.
\end{kaoboxA}
\vspace{-8pt}

\enlargethispage{4\baselineskip}
\begin{kaoboxA}[title=Gap 2: Verifiability of Outcomes]{gapmauve}{gapinkbg}
Software environments possess an elegant property that is easy to take for granted: every action has a natural \emph{termination signal} and an explicit \emph{success/failure judgment}.
Processes exit.
Commands return exit codes.
Errors print to \code{stderr}.
A coding agent runs a test and instantly knows \code{PASS} or \code{FAIL}, and on failure reads the stack trace to diagnose the cause.
This closed loop is infrastructure provided for free by the operating system and the language itself.

The physical world offers none of this.
A Vision-Language-Action (VLA) policy outputs a continuous action stream with no built-in ``done'' signal.
There is no exit code; the world does not report ``grasp succeeded.''
The robot closes its gripper. Did it grasp the cup, or did the cup slip? Did the gripper close on air, or on the rim?
A coding agent's ``test suite'' is free; a physical agent's must be constructed.
\end{kaoboxA}

What appears ``free'' to a coding agent is the accumulated product of decades of software engineering.
Formal grammars, type systems, and persistent file systems make the state of software readable; exit codes, stack traces, and test frameworks make the outcome of an action verifiable.
None of it was built for coding agents; they arrived to find the infrastructure ready, and flourished.
Physical environments carry no such inheritance; the missing infrastructure must be built.
Nor are the two gaps an arbitrary pair.
An agentic loop reaches its environment only through an interface with two directions: actions flow out, and information flows back.
The outbound half is the one robotics has spent decades building, which is why policies wrap readily into tools.
The inbound half carries exactly two signals: the state of the world, and a verdict on the last action, the same pair the classic agent--environment interface returns~\sidecite{sutton2018rlbook}.
Hence exactly two gaps, one for each missing signal.

We present \textsc{Thea}\sidenote{From the Greek \emph{thea}, ``sight; view.''}, a harness of embodied agents.
\thea inherits the architecture of coding agents: the same agentic loop, capabilities exposed as tools, context, skills, and memory.
On this foundation, it supplies the pieces the physical world is missing.
\emph{Scene Graph as Context} restores readability: a persistent, structured model of the scene that the language model can read and reason about as a coding agent reads source code.
\emph{Evaluation as Exit Codes} restores verifiability: it detects when an action should terminate, judges whether it succeeded, and on failure diagnoses the cause, closing the loop that the physical world otherwise leaves open.

\medskip
\noindent Through \thea, we demonstrate that with carefully adapted designs, the harness behind coding agents fits embodied agents well. It wires diverse foundation models, policies, and embodiments into a working whole.
The resulting system exhibits the following properties, as shown in Figure~\ref{fig:overview}:

\begin{itemize}[nosep]
  \item \textbf{Extensibility.} Capability is organized around tools. Adding a tool extends the agent, a new policy enters as one more tool, and complex behaviors emerge from their free composition rather than being hand-crafted.
  \item \textbf{Portability.} The model and the body are plug-and-play. A different model or a different embodiment requires no dedicated redesign, because nothing in the architecture is specific to either.
  \item \textbf{A bridge between user and world.} To the user, the agent is a natural interface. Instructions, questions, and explanations all flow through dialogue. To the world, it runs the classic perception--action loop autonomously, turn after turn. The agent bridges the two, translating the user's intention into fluent execution in the physical world.
\end{itemize}

\section{Why the Harness Works}
\label{sec:arithmetic}

The paradigm of coding agents transfers, in principle, to the physical world. Each robot capability becomes a callable tool, and a language model orchestrates them through an agentic loop~\sidecite{berman2026robotics}.
Our goal is to build the harness that makes this transfer work.
Yet the harness touches only orchestration, not the components themselves. The same policy, called through the harness, produces the same distribution of outcomes on any single attempt.
If no individual component becomes more likely to succeed, why should the system as a whole?
We formulate this question and analyze what factors govern the gain and what they imply for design.

Consider an $n$-step task\marginnote{For a household robot, a long-horizon task may require navigating across rooms and interacting with multiple objects, leading to a large $n$.} in which step $i$ succeeds with probability $p_i$.
Under open-loop execution the task succeeds only if every step does:
\begin{equation}
P_{\text{open}} = \prod_{i=1}^{n} p_i
\label{eq:open}
\end{equation}
Each factor below one compounds the loss; the product decays geometrically with $n$.

Now suppose the harness wraps each step in a detect-and-retry loop, allowing up to $k$ attempts per step.
After each attempt, an evaluator classifies the outcome as success or failure with accuracy $\alpha$: it returns the correct judgment with probability $\alpha$ and the wrong one with probability $1 - \alpha$.%
\marginnote{The four outcomes per attempt: true success correctly passed, $p_i\alpha$; true success misclassified and retried, $p_i(1{-}\alpha)$; true failure correctly detected and retried, $(1{-}p_i)\alpha$; true failure missed and passed through, $(1{-}p_i)(1{-}\alpha)$.}
A retry is triggered whenever the evaluator reports failure, whether correctly or not, with probability $r_i = p_i(1-\alpha) + (1-p_i)\alpha$.
We seek $p_i^*(\alpha, k)$, the probability that step $i$ produces a true success within $k$ attempts.
The step succeeds on attempt $j$ ($j = 1, \dots, k$) if all preceding $j{-}1$ attempts triggered retries (probability $r_i^{\,j-1}$) and the $j$-th attempt is a true success correctly passed (probability $p_i\alpha$).
These events are mutually exclusive, so:
\begin{equation}
p_i^{*}(\alpha, k)
\;=\; \sum_{j=1}^{k} r_i^{\,j-1} \cdot p_i\,\alpha
\;=\; p_i\,\alpha \cdot \frac{1 - r_i^{\,k}}{1 - r_i}
\label{eq:pstar}
\end{equation}
\noindent Substituting\marginnote[-1.5cm]{Eq.~\ref{eq:pstar} assumes each retry is an independent draw with the same $p_i$; we discuss this assumption further below.\\[4pt]
Special cases: at $\alpha {=} 1$, $r_i {=} 1{-}p_i$ and $p_i^* {=} 1 - (1{-}p_i)^k$ (perfect evaluator, failure compressed exponentially in $k$);\\[2pt]
at $\alpha {=} \tfrac{1}{2}$, $p_i^* {=} p_i(1 - 2^{-k}) \leq p_i$ (random guessing never exceeds the single-attempt rate);\\[2pt]
for sufficiently large $k$, $p_i^*$ approaches the ceiling $\frac{p_i}{(2p_i-1)+(1-p_i)/\alpha}$, monotonically increasing in $\alpha$, reaching~$1$ at $\alpha{=}1$.} into the product over all steps, the task reliability under the harness becomes:
\begin{equation}
P_{\text{harness}}
\;=\; \prod_{i=1}^{n} p_i^{*}(\alpha, k)
\;=\; \prod_{i=1}^{n} p_i\,\alpha \cdot \frac{1 - r_i^{\,k}}{1 - r_i}
\label{eq:harness}
\end{equation}
Each factor $p_i$ in the open-loop product is now scaled by a per-step multiplier $\alpha \cdot \frac{1 - r_i^{\,k}}{1 - r_i}$; even at moderate $\alpha$, the gains are substantial, as illustrated in Figure~\ref{fig:reliability}.\marginnote[0cm]{%
\includegraphics[width=\marginparwidth]{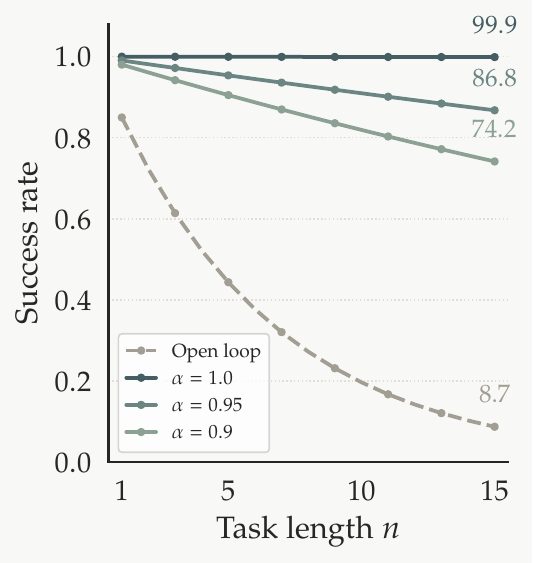}%
\captionsetup{labelsep=period}%
\captionof{figure}{Task success rate as a function of task length $n$ under open loop and under the harness at varying evaluation accuracy $\alpha$. Per-step success probability $p{=}0.85$, with up to $k{=}5$ retry attempts per step.}%
\label{fig:reliability}%
}

The formula makes explicit what the harness should provide: \emph{reliable evaluation} ($\alpha \to 1$).
Each additional retry yields geometrically less improvement, and the limit is directly capped by $\alpha$.
In software, $\alpha \approx 1$ is taken for granted; in the physical world, it must be actively constructed and maximized, making it the primary design challenge.
Also note that the formula rests on two simplifying assumptions, both conservative for a well-designed harness:
\begin{itemize}[nosep]
  \item \textbf{Retries within a step.}
  Eq.~\ref{eq:pstar} assumes each retry is an independent draw with the same $p_i$.
  Naively re-executing the same policy without adaptation may fail repeatedly, making retries \emph{worse} than independent draws.
  A well-designed harness, however, diagnoses the cause of failure and adapts the next attempt, for example by adjusting the robot's position or selecting a more suitable grasp policy, making retries progressively \emph{more} likely to succeed.
  \item \textbf{Fixed plan between steps.}
  The formulation holds the plan fixed (same $n$-step sequence under both open loop and harness execution).
  In practice, a robot may encounter unexpected situations that a fixed plan cannot anticipate: a door that was open during planning is now closed, or an object has been moved.
  The harness reads the world state before each step and re-plans adaptively, correcting course and effectively raising $p_i$ itself.
\end{itemize}
For a well-designed harness, both effects work in its favor; the formulation is therefore a lower bound on the true advantage.
The design of the harness presented in the following sections is informed by these considerations.

\section{System Architecture}
\label{sec:system}

This section is organized by a single question: what form does each component of a coding agent take in the physical world?
The answer has the following parts.
An \emph{agentic loop} (\S\ref{sec:agent-loop}) drives the agent. At each turn the \emph{model}, a vision-language model, reads the state of the world and issues a tool call for the harness to execute; we reserve \emph{policy} for the low-level controllers that tools invoke.
\emph{Context engineering} (\S\ref{sec:context}) determines what the model actually reads at each turn.
A \emph{tool protocol} (\S\ref{sec:tool-protocol}) defines how each capability is packaged so that the loop can execute it uniformly.
A \emph{skill system} (\S\ref{sec:skills}) injects domain knowledge into the context when the situation demands it.
A \emph{memory system} (\S\ref{sec:memory}) lets the agent accumulate what it learns, at every scope from a single task to the lifetime of a deployment.
Finally, \emph{safety} (\S\ref{sec:safety}) bounds what the agent may do at all.
Each component is inherited from the architecture of coding agents, and each is reshaped in some way by the demands of the physical world.
The components that the physical world requires, with no counterpart to inherit, follow in \S\ref{sec:gaps}.

\subsection{Agentic Loop}
\label{sec:agent-loop}

Listing~\ref{lst:loop} shows the agentic loop.%
\marginnote{%
  \footnotesize
  \textbf{Time units used throughout.}\\[3pt]
  \textbf{turn} --- one cycle of the loop: a model call plus the tool execution it requests.\\[3pt]
  \textbf{task} --- one user instruction handled to completion; it opens a run of the loop, spans as many turns as needed, and ends when the model returns no tool call.\\[3pt]
  \textbf{session} --- one continuous engagement with the user; tasks follow one another within it, and the accumulated messages carry across them (\S\ref{sec:context}).%
}
Each turn, the harness gathers the context, the model reads it and picks the next action, and the harness executes the tool call it names and appends the result; a response with no tool call ends the task.
This is the same loop that drives coding agents: the same accumulating messages, the same termination convention.
The control flow fits in these few lines.
\marginnote[3.2cm]{%
  \footnotesize
  \code{build\_context} --- assembles everything the model reads, by lifetime; expanded in \S\ref{sec:context} (Listing~\ref{lst:context}).\\[3pt]
  \code{update\_memory} --- consolidates the finished task's record into durable memory; the subject of \S\ref{sec:memory} (Listing~\ref{lst:consolidation}).%
}

\begin{lstlisting}[
  language=Python,
  float=tb,
  caption={Agentic loop.},
  label={lst:loop},
]
def agent_loop(instruction, session) -> str:
    """Perceive, decide, act, repeat until done."""
    messages = session.messages
    messages.append(instruction)

    while True:
        # perceive: gather what the model reads
        context = build_context(session)

        # decide: the model picks the next action
        response = llm(context)
        messages.append(response)
        if not response.tool_calls:  # done
            update_memory(session)
            return response.text

        # act: execute one tool call per turn
        call = response.tool_calls[0]
        messages.append(execute(call))
\end{lstlisting}

Everything the model reads is curated in \code{build\_context}, from durable instructions to physical evidence gathered fresh every turn.
Section~\ref{sec:context} unpacks this function, and the world state it draws on is the subject of \S\ref{sec:scene-graph}.

As in coding agents, every capability is exposed as a \emph{tool}. Navigation, manipulation, evaluation, and user interaction are all entries in the same flat \emph{tool registry}, and the loop has no special-cased logic for any of them.
The model issues exactly one \emph{tool call} per turn, because physical actions and their consequences are hard to predict.
This \emph{reactive} principle (execute one tool call, observe the outcome, then decide the next) lets the agent adapt to a world it cannot fully anticipate.
The absence of a tool call closes the task, and the final text reports its outcome.

Under this design, \emph{evaluation} is also a tool, but its invocation is not left to the model's discretion.
In coding agents, evaluation is invisible. A shell command returns an exit code and a stack trace, and the model reads the result like any other tool output.
The physical world provides no such signal, so \thea makes evaluation an explicit tool, and the harness triggers it structurally. A post-execution hook invokes the evaluator after every manipulation (\S\ref{sec:eval}).

\subsection{Context Engineering}
\label{sec:context}

\begin{figure}[!tb]
  \centering
  \includegraphics[width=\linewidth]{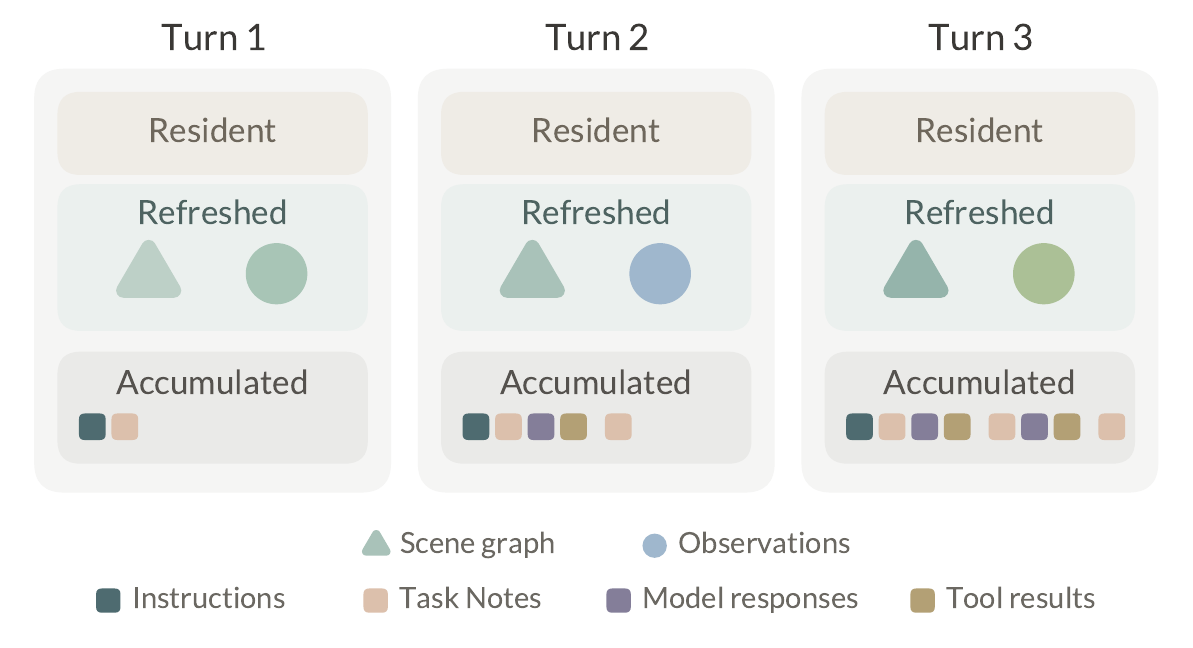}
  \refstepcounter{figure}
  \label{fig:context-engineering}
  \par\smallskip
  \parbox{\linewidth}{\raggedright\small\textbf{Figure~\thefigure.} Context evolution across ordinary turns. Resident context stays stable, refreshed context is replaced before each model decision, and accumulated context grows. Compaction and task-end expiry are not shown.\par}
\end{figure}

Context engineering curates what the model reads. The context window is a finite, attention-limited resource, and what fills it largely determines what the agent does~\sidecite{anthropic2025contextengineering}.
Unlike coding agents, which can often recover context by reading durable files and preserving a growing transcript, \thea must also refresh physical evidence whose validity changes with robot motion and external events.
Context engineering in \thea therefore assigns each model-visible input both a representational role and a lifetime, so stable operating knowledge, current physical evidence, and historical trace occupy distinct parts of the context window: resident, refreshed, and accumulated.
Listing~\ref{lst:context} unpacks \code{build\_context} from Listing~\ref{lst:loop}, Table~\ref{tab:context-lifetimes} lists the contents of each lifetime, and Figure~\ref{fig:context-engineering} shows how they evolve across turns.

Resident context holds the System Prompt, Memory, Embodiment Profile, and Tool Definitions.
Accumulated context holds Instructions, Task Notes, Model Responses, and Tool Results.
These two lifetimes behave just as they do in coding agents. Instructions stay, and the record grows.
Refreshed context holds the Scene Graph Brief and Observations, and is where the difference above lands.
Observations are the body's current sensor readings: camera images and clearance measurements, the LiDAR-detected distances to nearby obstacles.
They expire quickly, and their durable content is distilled into the scene graph, a holistic representation of the world, so the latest of each is all a decision needs.
Context-truncation experiments in robot manipulation point the same way. Models rely on the recent past far more than on a broad accumulated history~\sidecite{berman2026robotics}.
Appendix~\ref{app:context} expands each item to its full model-facing structure.

\marginnote{%
  \footnotesize
  Each lifetime has its own source: the resident blocks are loaded at task start, the refreshed values are pulled from the sensor side, and the accumulated record lives and grows in the session.%
}
\begin{lstlisting}[
  language=Python,
  float=tb,
  caption={Context assembly for one turn.},
  label={lst:context},
]
def build_context(session) -> Context:
    session.messages.append(task_notes())
    return Context(
        # resident: loaded at task start
        session.system_prompt, session.memory,
        session.profile, session.tool_schemas,
        # refreshed: pulled fresh, kept latest
        scene_graph.latest(), observe(),
        # accumulated: the growing record
        session.messages,
    )
\end{lstlisting}

\begin{table}[tb]
  \centering
  \small
  \begingroup
  \newdimen\contextlifetimeswidth
  \contextlifetimeswidth=.92\linewidth
  \setlength{\tabcolsep}{0pt}
  \setlength{\aboverulesep}{0.75ex}
  \setlength{\belowrulesep}{0.75ex}
  \refstepcounter{table}\label{tab:context-lifetimes}%
  \parbox{\contextlifetimeswidth}{\raggedright\small\textbf{Table~\thetable.} Context lifetimes for one turn: resident context stays stable, refreshed context is replaced before each decision, and accumulated context grows through ordinary turns. Compaction may replace older history, and Task Notes expire with the active task.\par}
  \par\smallskip
  \begin{tabular*}{\contextlifetimeswidth}{@{}p{0.18\contextlifetimeswidth}@{\extracolsep{\fill}}p{0.76\contextlifetimeswidth}@{}}
    \toprule
    \textbf{Lifetime} & \textbf{Contents} \\
    \midrule
    \textbf{Resident}
    &
    \begin{minipage}[t]{\linewidth}
      \raggedright
      \textbf{System Prompt}. The agent's role and operating rules.\\[0.25em]
      \textbf{Memory}. Durable memory from \code{MEMORY.md}: preferences, conventions, and lessons.\\[0.25em]
      \textbf{Embodiment Profile}. Description of the active embodiment: body, cameras, frames, and physical limits.\\[0.25em]
      \textbf{Tool Definitions}. \code{name}, \code{description} (with the tool's experience summary appended), and \code{inputSchema}.\strut
    \end{minipage}
    \\
    \midrule
    \textbf{Refreshed}
    &
    \begin{minipage}[t]{\linewidth}
      \raggedright
      \textbf{Scene Graph Brief}. Compact rendering of the current global world state.\\[0.25em]
      \textbf{Observations}. Camera images and clearance measurements.\strut
    \end{minipage}
    \\
    \midrule
    \textbf{Accumulated}
    &
    \begin{minipage}[t]{\linewidth}
      \raggedright
      \textbf{Instructions}. Messages from the user.\\[0.25em]
      \textbf{Task Notes}. Concise active-task working record.\\[0.25em]
      \textbf{Model Responses}. Model-generated messages, including text and previous tool calls.\\[0.25em]
      \textbf{Tool Results}. Return envelopes from executed tools.\strut
    \end{minipage}
    \\
    \bottomrule
  \end{tabular*}
  \endgroup
\end{table}

\paragraph{Resident.}
Resident context states the durable conditions for the next decision.
The System Prompt states the agent's role and its operating rules: one tool call per turn, which evidence source serves which purpose, and when to re-observe instead of trusting stale context.
Memory contributes durable knowledge carried across sessions, such as preferences, conventions, and lessons, while tool-specific experience is keyed separately (\S\ref{sec:memory}).
The Embodiment Profile gives the model the body's cameras, frames, and physical limits in the same place every time (\S\ref{sec:embodiment}).
Tool Definitions are the model-visible side of each tool's contract (\S\ref{sec:tool-protocol}); the post-condition never enters the model's context, and only the evaluator reads it (\S\ref{sec:eval}).
When a tool has accumulated experience, the harness appends a compact summary of it to that tool's description at task start (\S\ref{sec:memory}).

\paragraph{Refreshed.}
Refreshed context states what the harness currently claims about the world, at two scales.
The scene graph supplies the global scale: a structured summary of what is where, distilled from everything the robot has observed so far.
The graph itself lives in the harness; what enters the context is its compact rendering, the scene graph brief, regenerated with source and freshness metadata before each model call (\S\ref{sec:scene-graph}).
Observations supply the local scale: current camera images and clearance measurements.
These distances report the nearest obstacles forward, backward, left, and right for local motion decisions, not the semantic distance to a target.
The two complement each other. Observations are raw and instantaneous, valid for this decision only, while the scene graph is organized and persistent, yet what it asserts is always about the present.
Together they let the model ground every action in the world as it is, not as it was when some earlier tool ran.

\paragraph{Accumulated.}
Accumulated messages preserve what has happened within the session.
When a new instruction arrives inside the same session, the harness appends it to existing messages rather than opening a separate transcript.
Before each decision, the harness also appends Task Notes as a concise active-task working record.
Inside the loop, Model Responses add model-generated text and previous tool calls, and Tool Results add compact return envelopes from executed tools.
These entries help recovery by recording what was tried and what came back.
Compaction touches only this lifetime. As a conversation nears the model's context window, older messages are summarized into a shorter record, while resident and refreshed context are preserved~\sidecite{anthropic2025contextengineering}.

\subsection{Tool Protocol}
\label{sec:tool-protocol}

Every capability in \thea is a tool: a \code{name}, a \code{description}, and an \code{inputSchema}, the tool \emph{definition} of the Model Context Protocol~\sidecite{anthropic2024mcp}.
Before a call, this definition is everything the model knows about what a tool does, which is why coding agents treat this agent-computer interface with the same care as an interface built for people~\sidecite{yang2024sweagent,anthropic2024buildingagents,anthropic2025writingtools}.
What the physical world changes is how much this interface must specify.
The rest of this section unpacks the protocol: the contract a tool file carries, how it registers, the envelope every call returns, who executes it, and the hooks it passes through.

\paragraph{The contract.}
A coding tool rarely needs to explain when it will fail, but a physical tool wraps a policy with a success distribution, so its description states preconditions (\eg ``target within reach, gripper empty''), the character of the underlying policy, and what to try when it fails.
Each tool file also binds its own \emph{post-condition}, a few sentences stating what the world should look like if this particular call succeeded; at evaluation time the evaluator retrieves it by tool name, so the success criterion travels with the tool rather than with the prompt, and the model never sees it (\S\ref{sec:eval}).
In effect, each tool file carries the tool's full contract: the \code{inputSchema} says how to call it, the \code{description} says when, and the post-condition says how its outcome will be judged.
\newsavebox{\toolcontractbox}
\begin{lrbox}{\toolcontractbox}
\setlength{\fboxsep}{5pt}%
\colorbox{codebg}{\begin{minipage}{\dimexpr\marginparwidth-10pt\relax}
\ttfamily\scriptsize\raggedright
{\color{codekw}\bfseries def} pick\_up(\\
\hspace*{0.8em}object\_name: Annotated[str,\\
\hspace*{1.6em}Field(description={\color{codestring}"..."})],\\
\hspace*{0.8em}...,\\
) -> dict:\\
{\color{codestring}\hspace*{0.8em}"{}"{}"Run CaP-X pick-object;\\
\hspace*{0.8em}returns a run id.}\\[3pt]
{\color{codestring}\hspace*{0.8em}When to use: ...\\
\hspace*{0.8em}Readiness: ...\\
\hspace*{0.8em}Do not use: ...\\
\hspace*{0.8em}Result: ...\\
\hspace*{0.8em}"{}"{}"}\\[3pt]
{\color{codecomment}\itshape \# success criterion: fetched by\\
\# evaluator, never shown to model}\\
POST\_CONDITION = ({\color{codestring}"..."})
\end{minipage}}%
\end{lrbox}
\refstepcounter{lstlisting}\label{lst:schema}%
Listing~\ref{lst:schema} shows the shape of one such contract as deployed.%
\marginnote[-2.6cm]{%
  \usebox{\toolcontractbox}\\[4pt]
  {\scriptsize\textbf{Listing~\thelstlisting.} A deployed tool file (\code{pick\_up}), contents elided; it registers as the \code{Tool} of Listing~\ref{lst:tool}.}%
}

\begin{figure}[t]
  \centering
  \includegraphics[width=\textwidth]{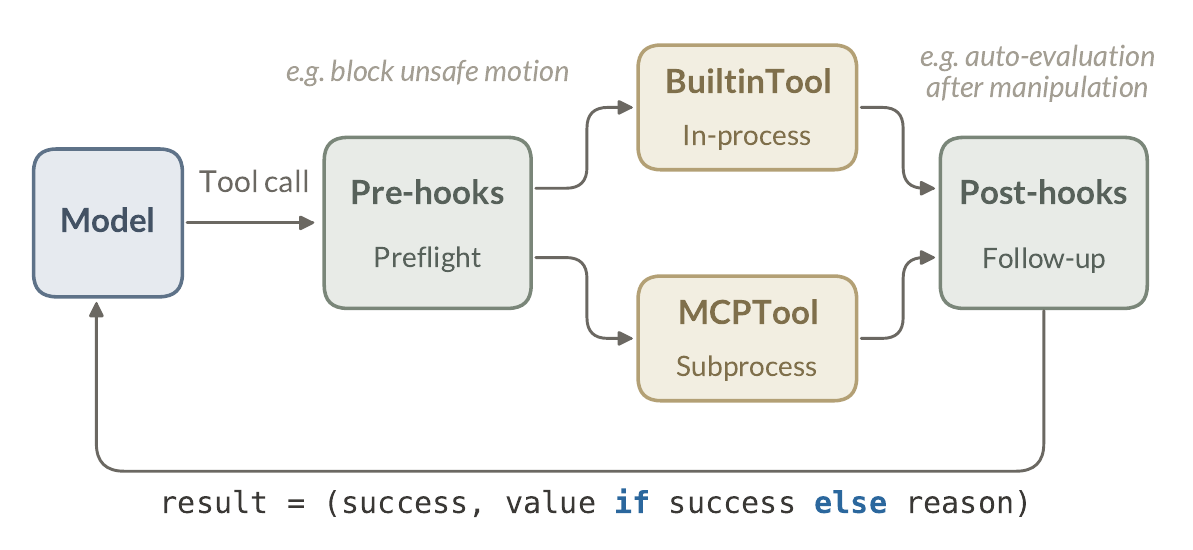}
  \refstepcounter{figure}
  \label{fig:tool-protocol}
  \smallskip
  \parbox{\textwidth}{\raggedright\small\textbf{Figure~\thefigure.} The model issues a tool call. Pre-hooks preflight the call, one backend executes it, and post-hooks trigger follow-up work before the result returns to the model.\par}
\end{figure}

\paragraph{Registration.}
A one-line registration, \code{server.tool()(pick\_up)}, turns the file of Listing~\ref{lst:schema} into the \code{Tool} of Listing~\ref{lst:tool}: the function name, the docstring, and the signature compile into the three fields of \code{schema} (\code{name}, \code{description}, \code{inputSchema}), and the body is what \code{call()} executes.
Listing~\ref{lst:tool} also shows the envelope every call returns; who executes it, and the hooks a call must pass through, are traced by Figure~\ref{fig:tool-protocol}.

\begin{lstlisting}[
  language=Python,
  float=tb,
  caption={The tool protocol: one interface and one result envelope for every tool.},
  label={lst:tool},
]
class Tool(Protocol):
    schema: dict  # the MCP tool definition:
                  # {name, description, inputSchema},
                  # introspected at registration from
                  # the file's signature and docstring
    def call(self, args: dict) -> dict:
        # executes the tool file's function body
        ...  #    {"success": True,  **value}
             # or {"success": False, "reason": reason}
\end{lstlisting}

\paragraph{The envelope.}
Every tool returns the same envelope: a \code{success} flag, a value when it succeeds, and a \code{reason} when it fails.
The value is, in the loop's vocabulary, the model's next observation; for a policy-backed tool it can be as thin as a run handle, an identifier for the ongoing policy execution (a \emph{run}), with the substantive evidence arriving when the post-hook brings in the evaluator (\S\ref{sec:eval}).
Failure never raises an exception; a failed call returns a \code{reason} such as ``the robot is too far from the target to grasp'', the closest thing a physical action has to \code{stderr}.
How verdicts and reasons are produced is the subject of \S\ref{sec:eval}; what the agent does with them is deliberately unscripted. Recoveries emerge from the model recombining the same tools, not from hand-designed recovery routines.

\paragraph{Execution.}
Who executes a call depends on its weight. Lightweight queries run in-process (\code{BuiltinTool} in Figure~\ref{fig:tool-protocol}), while heavyweight capabilities (\eg the navigation stack, the manipulation policies) each run in their own subprocess speaking the Model Context Protocol~\sidecite{anthropic2024mcp} (\code{MCPTool}).
The registry presents both as one flat list, and the model cannot tell them apart.
This indifference is what portability rests on. A policy backend can crash, restart, or be upgraded behind the same interface, and nothing above the protocol notices; the same subprocess boundary is what lets one harness drive backends written against three different robot stacks.%
\marginnote{%
  \footnotesize
  Table~\ref{tab:tools} (Appendix~\ref{app:tools}) lists the tool registry deployed on one embodiment.%
}

\paragraph{Hooks.}
A call never travels straight from the model to the tool (Figure~\ref{fig:tool-protocol}).
Each call passes through deterministic interception points that the harness owns and the model cannot skip: pre-execution hooks that preflight the call (and may rewrite or block it), and post-execution hooks that trigger follow-up work; an intercepted call returns through the ordinary result channel like any other failure.
The two that matter most are a safety check that reads fresh clearance measurements before any base motion (\S\ref{sec:safety}) and a hook that invokes the evaluator automatically after every manipulation (\S\ref{sec:eval}).
One discipline decides where each rule lives: rules about a single tool go into its description; rules the model must never be trusted to follow go into hooks; neither belongs in the system prompt, which stays short and tool-agnostic.

\FloatBarrier

\subsection{Skills}
\label{sec:skills}

Tools give the model capabilities; skills give it knowledge.
Concretely, a skill is a directory holding a \code{SKILL.md} file: YAML front matter carrying a \code{name} and a \code{description}, an instruction body in free-form markdown, and optionally bundled resources such as reference files or scripts.
Loading follows the progressive disclosure of coding agents~\sidecite{anthropic2025agentskills}, in three levels: the name and description of every registered skill stay resident in the context at a cost of tens of tokens each, the body enters the context through \code{load\_skill} only when the model judges that the task at hand matches a description, and bundled resources are read only if the instructions call for them.
Nothing in this mechanism is specific to robots; it transfers unchanged.

What the physical world sharpens is the boundary between skills and tools.
A tool is a contract: schema validation, safety checks, permissions, and evaluation all attach at the call boundary (\S\ref{sec:tool-protocol}).
A skill is advice: text the model reads and may weigh.
In a world with no sandbox and no undo, whatever acts must sit on the contract side, so a manipulation policy is always wrapped as a tool and never delivered as a skill that teaches the model how to invoke it; on the advice side the guarantees would have nothing to attach to.
Skills are left holding exactly what text is good at: knowledge.
Which knowledge is decided by elimination: rules about a single tool go into its description, behavior needed on every turn goes into the System Prompt (\S\ref{sec:context}), lessons the system accumulates on its own go into memory (\S\ref{sec:memory}), and what is left falls to skills.
In practice this ranges from the operating sequence of a particular appliance to the house rules of a particular workspace.
\newsavebox{\skillmdbox}
\begin{lrbox}{\skillmdbox}
\setlength{\fboxsep}{5pt}%
\colorbox{codebg}{\begin{minipage}{\dimexpr\marginparwidth-10pt\relax}
\ttfamily\scriptsize\raggedright
{\color{codecomment}-{}-{}-}\\
{\color{codekw}\bfseries name}: tidy-workspace\\
{\color{codekw}\bfseries description}: Rules for tidying\\
\hspace*{1em}a desk: what to remove, what\\
\hspace*{1em}to keep, and how to report.\\
{\color{codecomment}-{}-{}-}\\
{\bfseries\color{codebuiltin}\#\# Working order}\\
- ...\\
{\bfseries\color{codebuiltin}\#\# What to remove / what to keep}\\
- ...\\
{\bfseries\color{codebuiltin}\#\# Wrap-up}\\
- ...
\end{minipage}}%
\end{lrbox}
\refstepcounter{lstlisting}\label{lst:skill}%
Listing~\ref{lst:skill} shows the shape of one deployed in \thea: a page of boundary judgments for tidying a desk, knowledge at the level of convention rather than of operation.%
\marginnote[-8cm]{%
  \usebox{\skillmdbox}\\[4pt]
  {\scriptsize\textbf{Listing~\thelstlisting.} A skill is a markdown file: the deployed \code{tidy-workspace/SKILL.md}, contents elided.}%
}
As a skill is only text, the skill system extends by editing a file: a new appliance, a new house rule, with no retraining and no code change.

\subsection{Memory}
\label{sec:memory}

\newsavebox{\memoryartifactsbox}
\begin{lrbox}{\memoryartifactsbox}
\setlength{\fboxsep}{5pt}%
\colorbox{codebg}{\begin{minipage}{\dimexpr\marginparwidth-10pt\relax}
\ttfamily\scriptsize\raggedright
{\bfseries\color{codebuiltin}Task Notes}\\
{\color{codekw}\bfseries Task summary}:\\
- Goal: current instruction\\
- Current phase: ...\\
{\color{codekw}\bfseries Timeline}:\\
- task\_start: ...\\[3pt]
{\bfseries\color{codebuiltin}Memory}\\
- [user-preference] User gives robot tasks in Chinese.\\[3pt]
{\bfseries\color{codebuiltin}Tool Experience}\\
\code{open\_drawer}\\
{\color{codekw}\bfseries Success}:\\
- Lower drawer pulled outward\\
\hspace*{0.8em}with a visible gap.\\
{\color{codekw}\bfseries Failure}:\\
- If distance is reported or no gap\\
\hspace*{0.8em}appears, move forward before retrying.
\end{minipage}}%
\end{lrbox}
Memory in \thea is a lifecycle rather than a single store.
\paragraph{Task Notes.}
Inside a running task, the agent first needs a notepad: a place that keeps what has been done and what remains, so that a long task never depends on the model re-deriving its own progress~\sidecite[-1.5cm]{anthropic2025contextengineering}.
In \thea this notepad is Task Notes, a concise record that preserves continuity inside one task.
A task begins when the user gives an instruction and ends when the model returns no tool call.
At task start, the harness resets Task Notes.
After each tool result, it appends a brief event.
The notes carry the current goal, phase, and timeline.
They are the working record for the active task, not a session transcript and not live world state.

\begin{figure}[!t]
  \centering
  \includegraphics[width=\textwidth]{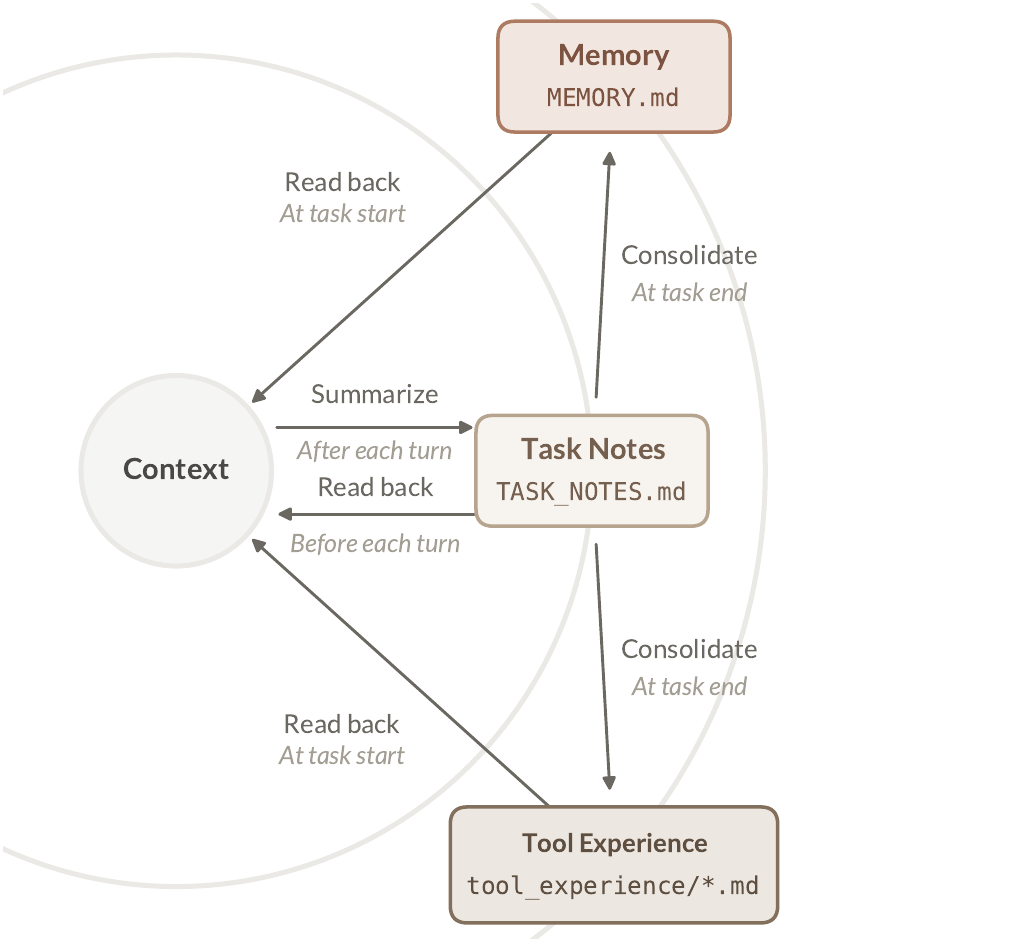}
  \refstepcounter{figure}
  \label{fig:memory}
  \par\smallskip
  \parbox{\textwidth}{\raggedright\small\textbf{Figure~\thefigure.} Memory lifecycle. Each store cycles around the Context hub at its own period. Within a task, the harness summarizes each turn's events into Task Notes and reads the snapshot back into the accumulated messages before each turn. At task end, consolidation writes accepted cross-task entries to \code{MEMORY.md} and accepted tool-scoped lessons to that tool's experience file. At task start, both are read back into resident context, Memory directly and tool experience as a summary inside each tool's description.\par}
\end{figure}

\paragraph{Durable stores.}
Beyond a single task, the agent needs durable memory.%
\refstepcounter{lstlisting}\label{lst:memory-artifacts}%
\marginnote[-4cm]{%
  \usebox{\memoryartifactsbox}\\[4pt]
  {\scriptsize\textbf{Listing~\thelstlisting.} Abridged memory artifacts: a Task Notes shape, a Memory entry, and success/failure tool experience.}%
}
Some of what it learns is general: who the user is and how things are usually done.
Some is specific to one tool: how it succeeds and how it fails, lessons that make the next call better.
Coding agents have little of the tool-specific kind: \code{grep} does exactly what its description says, every call, so the tool itself leaves nothing to learn.
A tool that wraps a policy is different. Its description promises only a tendency, and where it actually succeeds or fails must be learned from use~\sidecite[2.0cm]{qu2025draft} (\S\ref{sec:tool-protocol}).
\thea therefore keeps two durable stores: \code{MEMORY.md}, the same cross-session memory file a coding agent keeps, and \code{tool\_experience/}, one file per tool.
Both are written at task end, when the harness consolidates the completed Task Notes, splitting entries by where they will return to the model: cross-task knowledge to Memory, tool-scoped lessons to that tool's experience file.
The consolidation call scores each candidate, and the harness writes only those above a confidence threshold.
Evaluator verdicts (\S\ref{sec:eval}) enter this pipeline only as evidence in the trace, never as direct writes. A single verdict is a hypothesis, and whether the agent actually recovered from it shows only in the completed trajectory.
Listing~\ref{lst:consolidation} shows the consolidation step, and Figure~\ref{fig:memory} draws the full cycle.

\begin{lstlisting}[
  language=Python,
  float=tb,
  caption={Memory consolidation, run once at task end.},
  label={lst:consolidation},
]
def update_memory(session) -> None:
    """Consolidate a finished task into durable memory."""
    # one model call reads the completed Task Notes and
    # splits them: cross-task knowledge vs tool lessons
    memory, lessons = llm(CONSOLIDATE, session.task_notes)
    append_memory("MEMORY.md", memory)  # user preferences
    for lesson in lessons:              # how a grasp fails
        path = f"tool_experience/{lesson.tool}.md"
        append_memory(path, lesson.entry)
\end{lstlisting}

\paragraph{Read paths.}
The three read-back arrows differ in what the return buys.
Task Notes expire with the task; until then, each snapshot keeps the model current on its own progress.
Memory returns as resident context, so durable knowledge is visible on every decision without an extra tool call.
Tool experience is keyed by tool identity rather than by session or user; the harness appends a summary of each tool's file to that tool's description, so the lessons sit beside its preconditions, failure modes, and recovery hints, exactly where the model weighs whether to call it.
Description quality largely determines call quality~\sidecite{anthropic2025writingtools}, so the memory lifecycle does more than remember. What it writes and reloads is, in effect, the interface between the model and its tools.
The agent improves its own interface with use.

\FloatBarrier

\subsection{Safety}
\label{sec:safety}

Coding agents secure themselves with permission rules and an operating-system sandbox~\sidecite{anthropic2025sandboxing}: the rules tell the agent what it \emph{should} do, with denials enforced below the model, and the sandbox bounds what it \emph{can} do.
Neither is available in the physical world. No sandbox contains a physical action (simulation covers part of this need, but not the deployed world), and many actions cannot be undone.
Safety must therefore be enforced before the action.

\thea builds safety into the machinery rather than into the model's behavior~\sidecite{ahn2024autort}. Deterministic checks live in the hooks and the execution pipeline of \S\ref{sec:tool-protocol}, and the model can neither skip nor persuade them.
The main one is a \emph{safety filter} on base motion. Before the model decides, the filter puts four-direction clearance measurements into the refreshed context, so the model plans with nearby obstacles in view. Before the robot moves, its hook obtains a fresh reading, blocks navigation when no immediate direction is admissible, and clamps direct translations to the admissible distance.
The loop itself is conservative: physical actions run one at a time (\S\ref{sec:agent-loop}), a failure budget halts the loop instead of letting it run away, and when unsure the agent can stop and ask the user (\code{query\_user}, \S\ref{sec:user-in-loop}).
Safety is one of the deepest differences between coding agents and embodied agents in the physical world; \thea makes an initial attempt, and much remains open: force limits, safety around people, and deployment beyond the lab.

\subsection{User Interaction}
\label{sec:user-in-loop}

Coding agents keep the user reachable throughout a task. The model can ask a blocking clarification question or surface progress, and autonomy is defined with structured returns to the human for information or judgement~\sidecite{anthropic2024buildingagents}.
\thea carries this over as two tools whose endpoint is a person.
\code{query\_user} asks and waits: for a choice among lookalike targets, for confirmation of a borderline action, for permission before touching something personal.
\code{notify\_user} tells without waiting: that a long action is starting, that something unexpected happened, that the task is beyond what the robot can do.
When to call either is a tool choice like any other in the agentic loop (\S\ref{sec:agent-loop}).

In the harness, the two tools compose with everything else.
Because they are ordinary tool calls, everything that shapes tool choice shapes them too. A question comes as a last resort rather than a reflex, after the scene graph and the current observations have been consulted, and it can attach the graph's candidate refs and views so the answer returns as evidence the rest of the task can ground on (\S\ref{sec:scene-graph-interface}).
A skill can state when a notification is warranted for its task, and Memory keeps settled answers as standing preferences so the same question is not asked twice.
Keeping the user in the loop is what keeps an embodied agent adaptive to the unknown and the unexpected: it neither guesses nor fails silently.

\section{Bridging the Gaps}
\label{sec:gaps}

The previous section carries the harness across layer by layer. Every layer has a digital counterpart to start from, and the work is adaptation.
This section builds what has no counterpart.
Gaps 1 and 2 (\S\ref{sec:intro}) name the two absences that break the loop itself: the physical world is neither readable nor verifiable by construction, so the infrastructure a coding agent inherits for free must here be built.
The scene graph restores readability, giving the loop a world it can reference (\S\ref{sec:scene-graph}).
The evaluator restores verifiability, approximating the exit codes the world does not issue (\S\ref{sec:eval}).
A third absence has no named gap because coding agents never meet it. They have no body.
An embodied agent does, and what it can sense, reach, and do depends on which body it runs on.
The Embodiment Profile describes the body and the boundaries of its capabilities (\S\ref{sec:embodiment}).

\subsection{Scene Graph as Context}
\label{sec:scene-graph}

We restore world readability through two linked layers (Figure~\ref{fig:scene-graph-world-state}).
An object-centric scene graph maintains a persistent, symbolic world state across turns.
A decision-facing interface carries that state into the model's context: a compact scene graph brief by default, with queries keyed by object ref when more detail is needed.
\S\ref{sec:scene-graph-state} defines the world-state representation and its update process.
\S\ref{sec:scene-graph-interface} explains how the model reads that state, confirms an object ref, and carries the ref into tool calls.

\subsubsection{Persistent World State}
\label{sec:scene-graph-state}

A coding agent inherits a named and searchable environment. It can search for a symbol, read the path that comes back, and hand the same path to its next tool call.
A physical agent starts with none of this.
Its visual and depth observations describe one view at one moment, robot state and tool feedback arrive on separate channels, and none of these signals organizes the world into entities the loop can reference again.
Yet to continue a task, the loop must know which entity it is pursuing, where that entity was last observed, and how much to trust that evidence.
We therefore turn fleeting signals into persistent symbols. An object-centric scene graph ties a stable ref to each object's coarse position and freshness, turn after turn~\sidecite[-1.5cm]{gu2024conceptgraphs,rana2023sayplan,yoneda2024statler}.

\paragraph{Graph structure.}
The scene graph organizes world state into nodes, edges, and graph metadata, following 3D scene-graph representations that bind semantic entities to spatial structure~\sidecite{armeni2019scenegraph}.
An object node carries a ref such as \code{cup\_2}, which the harness derives from the node's \code{label} and \code{id}; behind the ref, the node stores a coarse 3D bounding box, confidence, freshness, and a link to visual evidence.
A container (\eg a cabinet or a drawer) keeps its state and tracked contents on the same node, and a robot node tracks pose and holding state.
Edges encode \code{on}, \code{inside}, \code{holding}, and \code{near} relations among object, container, and robot nodes; a \code{near} edge, for example, carries the measured distance between its endpoints.
Graph metadata records provenance, coordinate frame, and update time.
Listing~\ref{lst:scene-graph-node} shows one recorded object node.

\newsavebox{\scenegraphnodebox}
\begin{lrbox}{\scenegraphnodebox}
\setlength{\fboxsep}{5pt}%
\colorbox{codebg}{\begin{minipage}{\dimexpr\marginparwidth-10pt\relax}
\ttfamily\scriptsize\raggedright
{\bfseries\color{codebuiltin}Recorded object node}\\[2pt]
\{\\
\hspace*{0.8em}{\color{codekw}"id"}: 2,\\
\hspace*{0.8em}{\color{codekw}"label"}: {\color{codestring}"cup"},\\
\hspace*{0.8em}{\color{codekw}"center"}: [3.91, -0.18, 1.08],\\
\hspace*{0.8em}{\color{codekw}"extent"}: [0.099,\\
\hspace*{1.6em}0.133, 0.096],\\
\hspace*{0.8em}{\color{codekw}"confidence"}: 1.0,\\
\hspace*{0.8em}{\color{codekw}"last\_seen\_time"}:\\
\hspace*{1.6em}1778162862.957,\\
\hspace*{0.8em}{\color{codekw}"image\_path"}:\\
\hspace*{1.6em}{\color{codestring}".../2.jpg"}\\
\}\\[2pt]
{\color{codecomment}derived ref: cup\_2}
\end{minipage}}%
\end{lrbox}
\refstepcounter{lstlisting}\label{lst:scene-graph-node}%
\marginnote[-1.8cm]{%
  \usebox{\scenegraphnodebox}\\[4pt]
  {\scriptsize\textbf{Listing~\thelstlisting.} A recorded object node, abridged.}%
}

\paragraph{Graph maintenance.}
A persistent picture must be kept true, and two kinds of change reach it: what the robot sees, and what the robot does.
The perception backend supplies observation-derived state from visual and depth evidence, refreshing object nodes, coarse positions, confidence, freshness, and spatial relations, echoing dynamic scene-graph construction and structured representations used in real-world ObjectNav systems~\sidecite{rosinol2020dynamic,takmaz2023openmask3d,zhu2026sysnav}.
Execution-derived updates record what physical actions change, such as grasping an object or opening a container; action-conditioned scene graphs likewise use interactions to reveal previously hidden structure~\sidecite{jiang2025roboexp}.
In our harness, this input is gated. After a tool finishes, the evaluator checks its post-condition (\S\ref{sec:eval}), and only a confirmed outcome may edit the graph.
A confirmed pickup, for example, marks the robot as holding the object and removes it from where it lay; a confirmed cabinet opening records that container as open.
The harness resolves both inputs against object refs and merges them into a single graph.
Together they keep the scene graph an up-to-date picture of the world.

\begingroup
\widefloatsetup
\begin{figure}[!t]
  \centering
  \makebox[\textwidth][l]{%
    \includegraphics[width=\dimexpr\textwidth+\marginparsep+\marginparwidth\relax]{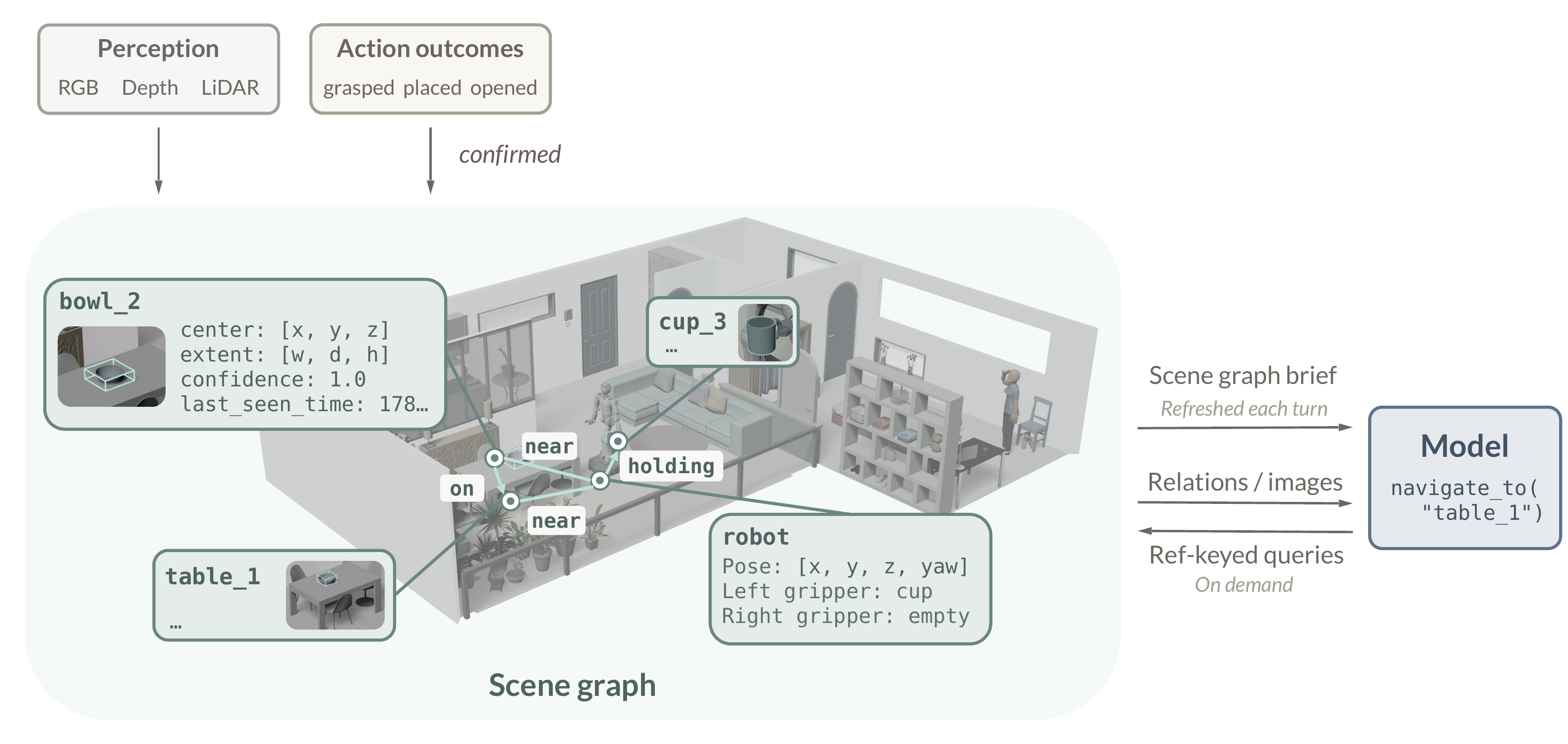}%
  }
  \refstepcounter{figure}
  \label{fig:scene-graph-world-state}
  \smallskip
  \makebox[\textwidth][l]{%
    \parbox{\dimexpr\textwidth+\marginparsep+\marginparwidth\relax}{\raggedright\small\textbf{Figure~\thefigure.} Maintaining and reading the world state.\sidenote[][1.6cm]{Scene layout designed with Sweet Home~3D. Includes 3D models and textures distributed under free licenses.} Observation-derived updates and evaluator-confirmed action outcomes are resolved against object refs and merged into one graph, carried into the next turn. The scene graph brief is pushed into the refreshed context every turn. Ref-keyed queries fetch omitted evidence on demand.\par}%
  }
\end{figure}
\endgroup

\subsubsection{Accessing the World State}
\label{sec:scene-graph-interface}

\paragraph{Scene graph brief.}
The full graph is too much to hand the model every turn. The context window is a finite, attention-limited resource (\S\ref{sec:context}), and most graph detail is irrelevant to the next decision.
The harness therefore renders a scene graph brief into the refreshed context, with frequently needed information present by default and the rest fetched on demand.
Nearly every decision must know which objects the world offers, where each one lies, and how far to trust that evidence, so the brief lists every object ref with its coarse position, confidence, and freshness.
The robot's own pose and holding state condition every action choice, and tracked container contents keep known but unseen objects addressable, so both enter the brief.
Detail that only particular decisions touch, such as an object's extent, its relation edges, and its stored images, stays in the graph and is retrievable by ref.
The brief is compact because it selects fields rather than pruning nodes; the boundary need not be exact, since anything omitted costs one extra query, not a failure.
The harness regenerates the brief before each call while preserving the graph across calls, so the model reads a current global picture without reconstructing world state from accumulated messages~\sidecite{anthropic2025contextengineering}.
Listing~\ref{lst:scene-graph-brief} shows an abridged brief.

\newsavebox{\scenegraphbriefbox}
\begin{lrbox}{\scenegraphbriefbox}
\setlength{\fboxsep}{5pt}%
\colorbox{codebg}{\begin{minipage}{\dimexpr\marginparwidth-10pt\relax}
\ttfamily\scriptsize\raggedright
Scene graph brief:\\
\hspace*{0.8em}objects=10;\\
\hspace*{0.8em}stamp=1778162862.957\\
Robot: pose=[0,0,0,0],\\
\hspace*{0.8em}gripper empty\\
Observed objects:\\
\hspace*{0.8em}- cup\_2 at (3.91,-0.18,1.08);\\
\hspace*{1.6em}conf=1\\
\hspace*{0.8em}- laptop\_5 at (5.21,1.18,1.29);\\
\hspace*{1.6em}conf=1\\
\hspace*{0.8em}...
\end{minipage}}%
\end{lrbox}
\refstepcounter{lstlisting}\label{lst:scene-graph-brief}%
\marginnote[0.2cm]{%
  \usebox{\scenegraphbriefbox}\\[4pt]
  {\scriptsize\textbf{Listing~\thelstlisting.} Scene graph brief rendered before a decision.}%
}

\paragraph{Ref-keyed queries.}
What the brief omits, the model fetches itself. A small set of query tools reads the full graph on demand, keyed by ref, so every answer attaches to an entity the graph already names rather than introducing a new one.
\code{get\_object\_relations} returns the typed relations of a ref, and \code{get\_image} returns the visual evidence stored on its node.
For example, the graph names objects at the class level, so when the user asks for ``the red cup'' and the brief lists three cups, no ref settles the instruction by name; the model calls \code{get\_image} on the candidates and reads the color from their stored views, evidence it needs for this one decision and not as standing context.
A spatial cue such as ``the cup by the desk'' resolves similarly through \code{get\_object\_relations}.
The design ends at the tools; what follows is the model's own behavior.
It chooses which query to issue and when, and asks through \code{query\_user} when no gathered evidence settles what the user means.
It treats an empty lookup as absence of evidence, not absence of the object, and obtains a fresh observation or inspects a likely container before concluding that the object is not there.
Composing these moves, the model resolves a vague phrase into a symbolic instance and hands its ref to the next action tool, \code{navigate\_to(target="cup\_2")}; the binding keeps naming the same entity through evaluation and retry until new evidence breaks it.

A confirmed ref settles which entity the robot pursues and where it approaches; it does not promise that manipulation is ready from the current pose.
That judgment falls to later evidence. Near the object, a fresh observation grounds visibility and alignment, and after each action the evaluator rules on the outcome (\S\ref{sec:eval}).
The scene graph supplies identity and coarse location, nothing more. It makes the world readable and addressable without becoming a readiness or success oracle.

\subsection{Evaluation as Exit Codes}
\label{sec:eval}

In the paradigm of coding agents, the environment itself tells the agent
how its last action went: commands return exit codes, tests fail with stack
traces, and the agent debugs against these signals~\sidecite{anthropic2026eval}.
The physical world provides no such infrastructure. A robot that has just
attempted a grasp receives no signal telling it whether the task is still in
progress, has succeeded, or has failed, let alone why: a slipped grip and a
collision both go unreported.
Supplying this missing verdict is neither optional nor easy.
The reliability analysis in \S\ref{sec:arithmetic} shows that evaluator accuracy $\alpha$ directly bounds the end-to-end task success rate: a harness is only as reliable as its judge.
The difficulty also inverts across worlds. In coding agents a single step judges itself, and the hard questions begin only at the level of whole trajectories; in the physical world, judging even one action is already the open problem.

\begin{figure}[t]
  \centering
  \refstepcounter{figure}
  \label{fig:evaluation_harness}
  \includegraphics[width=\linewidth]{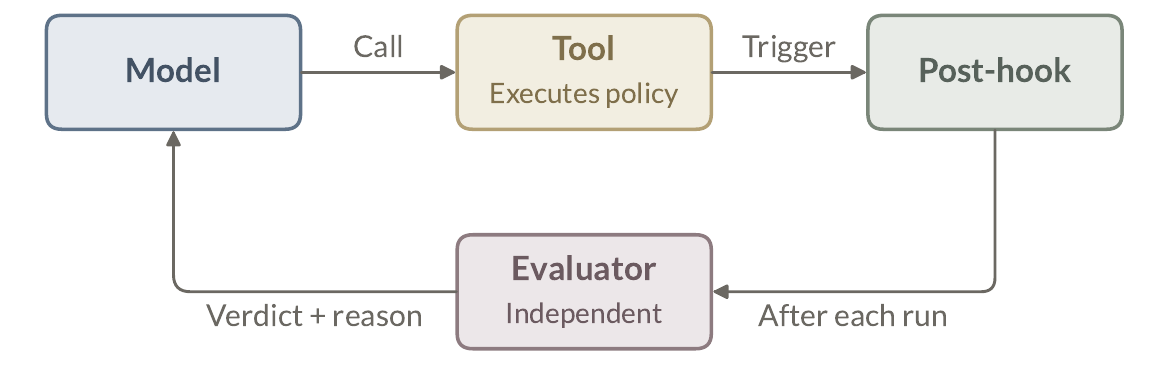}
  \par\smallskip
  \parbox{\textwidth}{\raggedright\small\textbf{Figure~\thefigure.} The harness triggers the evaluator structurally after every manipulation. The evaluator reads only the current observation and the per-tool post-condition, then returns a three-state verdict with a reason to the model.\par}
\end{figure}

\thea therefore makes evaluation an explicit module of the harness. An evaluator judges each action's outcome and returns the verdict with a structured failure reason, the exit code and the stack trace that the world never provides.
Designing it comes down to three questions, answered in turn below and illustrated in Figure~\ref{fig:evaluation_harness}: \emph{when} to judge, since not even the end of an action announces itself; \emph{who} judges, since a self-report of success cannot be the verdict; and \emph{what} the verdict carries, since a bare boolean gives replanning nothing to work with.

\paragraph{Timing.}
When to judge depends on the family of the policy behind the tool.
For end-to-end models~\sidecite{black2025pi0, black2025pi05}, no signal marks completion, so the evaluator judges the run online, at each action-segment boundary: \code{process} means the run should continue, \code{success} that the post-condition has been satisfied, and \code{failure} that execution should stop, with a step budget as the backstop.
For Coding-as-Policy~\sidecite{fu2026capx} approaches, the generated program terminates on its own and reports its progress through control flow, so the same contract degenerates to the two terminal states.

\paragraph{Independence.}
Who judges is settled structurally. The harness places the evaluator as an independent component after tool execution,
triggered by a post-hook rather than by the model's explicit invocation.
Its inputs are restricted to two things: the current observation, and the per-tool post-condition (what the world should look like after a successful call)~\sidecite{meyer1992dbc}.
It reads neither the model's reasoning nor its self-report, the reasoning-blind discipline Anthropic applies to its own action classifier~\sidecite{anthropic2026automode}. A model grading its own work is an unreliable judge~\sidecite{huang2024selfcorrect}, and a biased one~\sidecite{panickssery2024favor}.

\paragraph{The verdict.}
What the verdict carries determines what recovery can follow.
A bare status already closes the loop, but only blindly: the agent can retry, yet cannot adapt.
For example, if a grasping action fails, the agent needs to know whether the failure was caused by failing to reach the target, missing the object, object occlusion, or dropping the object after grasping; each cause calls for a different recovery.
Beyond the status, the evaluator therefore returns evidence and the failure reason; Listing~\ref{lst:eval-contract} shows the contract returned for one such failed grasp.
The division is deliberate~\sidecite[6.5cm]{liu2023reflect,guo2024doremi}. The evaluator only judges the outcome and explains the failure, and what to do next is left to the agent, which holds the full context the evaluator never sees.
Downstream, a \code{success} gates the scene graph update after the action (\S\ref{sec:scene-graph}), and the failure reasons feed the tool experience consolidated at task end (\S\ref{sec:memory}).

\begin{marginlisting}[-1.8cm]
  \begin{lstlisting}[language=json, basicstyle=\ttfamily\scriptsize, numbers=none, xleftmargin=0pt, framexleftmargin=0pt, framesep=5pt, caption={The evaluator contract for one failed grasp.}, label={lst:eval-contract}]
  {
    "status": "failure",
    "evidence": [
      "the bottle remains on the table",
      "the gripper closed without lifting it"
    ],
    "failure_reason": "the robot stopped too far from the bottle to grasp it"
  }
  \end{lstlisting}
  \end{marginlisting}

\subsection{Embodiment Profile}
\label{sec:embodiment}

A coding agent never needs to be told about a body.
It works through the shell and the file system, an interface that is the same everywhere and, when in doubt, can simply be asked.
A robot's body takes no such questions, and nothing about it is standard. Morphology, sensing geometry, mobility, and reach vary across platforms, and jointly determine what evidence is available, how spatial measurements should be interpreted, and which actions are feasible.
The Embodiment Profile states these body-specific facts explicitly, as a compact, replaceable document.
At deployment, the harness loads the profile selected for the active embodiment and retains it throughout the session.
Replacing the profile changes the body-specific context as a unit while task instructions retain the same form. This is what makes one harness portable across embodiments.
Localizing embodiment knowledge in one document keeps it consistent across decisions and makes adaptation easier to inspect, maintain, and validate.

The Embodiment Profile organizes stable body-specific information into three
complementary sections, summarized in Table~\ref{tab:embodiment-profile}.

\begin{center}
  \begingroup
  \small
  \setlength{\tabcolsep}{4pt}
  \renewcommand{\arraystretch}{1.22}
  \refstepcounter{table}
  \label{tab:embodiment-profile}
  \begin{minipage}{\textwidth}
  \parbox{\textwidth}{\raggedright\small\textbf{Table~\thetable.} The three sections of the Embodiment Profile, their stored items, and their role in model decisions.\par}
  \par\smallskip
  \begin{tabular}{@{}p{0.24\textwidth}p{0.34\textwidth}p{0.33\textwidth}@{}}
    \toprule
    \raggedright\textbf{Profile item}
    & \raggedright\textbf{Stored information}
    & \raggedright\textbf{Decision support} \tabularnewline
    \midrule
    \multicolumn{3}{@{}l}{\textbf{Operational Envelope}} \tabularnewline
    \addlinespace[0.25em]
    \raggedright Base Footprint
    & \raggedright Radius or dimensions of the body's ground footprint.
    & \raggedright Relates available space to the area occupied by the body. \tabularnewline
    \raggedright Base Mobility
    & \raggedright Supported base translations and rotations.
    & \raggedright Rules out motion commands that the base cannot execute. \tabularnewline
    \raggedright Reachable Workspace
    & \raggedright End-effector reach and operating height.
    & \raggedright Determines whether an object or location is physically reachable. \tabularnewline
    \midrule
    \multicolumn{3}{@{}l}{\textbf{Perception Configuration}} \tabularnewline
    \addlinespace[0.25em]
    \raggedright Sensor Modalities
    & \raggedright Types of evidence available to the model.
    & \raggedright Identifies whether a decision can use images or distance evidence. \tabularnewline
    \raggedright Model-Visible Views
    & \raggedright Named mounted camera views exposed to the model.
    & \raggedright Identifies the views that the model can request. \tabularnewline
    \midrule
    \multicolumn{3}{@{}l}{\textbf{Base-Relative Positions}} \tabularnewline
    \addlinespace[0.25em]
    \raggedright Camera Positions
    & \raggedright Fixed camera positions relative to the base center.
    & \raggedright Situates visual evidence on the active body. \tabularnewline
    \raggedright Initial Gripper Positions
    & \raggedright Gripper positions in the initial posture relative to the base center.
    & \raggedright Situates the manipulation interfaces on the active body. \tabularnewline
    \bottomrule
  \end{tabular}
  \end{minipage}
  \endgroup
\end{center}

Operational Envelope records the body's stable limits for motion and
manipulation. The base is whatever carries the body, a wheeled chassis or a
humanoid's legs, and its entries record what the base can do, not how it does
it. Perception Configuration describes the evidence made available to the
model, such as color images and clearance measurements. Base-Relative
Positions records where sensing and manipulation interfaces are located
relative to the base center. Taken together, Operational Envelope constrains
feasible actions, Perception Configuration identifies the evidence available
for a decision, and Base-Relative Positions situates that evidence and the
manipulation interfaces on the active body.

\section{Experiments}
\label{sec:experiments}

We evaluate \thea on real robots in real-world environments.
The experiments explore three questions.
The first is how the full harness compares with alternative execution architectures as task complexity increases.
The second is how reliably the evaluator, on which closed-loop execution depends, judges the state of an ongoing task.
The third is what capabilities emerge from the composition of tools in complete deployments across different embodiments.

\subsection{Task Setup}
\label{sec:exp-setup}

Our experiments span three robots with distinct embodiments. The main quantitative experiments run on the Astribot S1, a wheeled dual-arm humanoid that combines an actuated head and torso, an omnidirectional mobile base, and two gripper end effectors, 25 degrees of freedom in total. We additionally explore \thea on AgileX Cobot Magic, which pairs two arms with a mobile base, and on Unitree G1, which carries BrainCo Revo 2 dexterous hands with six degrees of freedom per hand. These deployments test whether the same harness can compose capabilities exposed by different physical interfaces.

\begin{table}[h]
  \centering
  \refstepcounter{table}\label{tab:task-hierarchy}%
  \parbox{0.9\linewidth}{\raggedright\small\textbf{Table~\thetable.} Real-world tasks of increasing difficulty.\par}
  \par\smallskip
  \small
  \begin{tabular}{@{}m{0.08\linewidth}>{\raggedright\arraybackslash}m{0.34\linewidth}m{0.48\linewidth}@{}}
    \toprule
    Level & Task & Task description \\
    \midrule
    L1 & Short-horizon manipulation & Pick up one tabletop object and place it into a basket. \\
    \addlinespace[6pt]
    L2 & Long-horizon manipulation & Pick up all tabletop objects and place them into a basket. \\
    \addlinespace[6pt]
    L3 & Navigation and manipulation & Navigate to one table, retrieve the instructed object, and deliver it to a target table. \\
    \bottomrule
  \end{tabular}
\end{table}

To evaluate \thea and the baselines across tasks of increasing difficulty, we organize the experiments into three levels, L1--L3, as summarized in Table~\ref{tab:task-hierarchy}. L1 is a simple pick-and-place task that isolates a single manipulation cycle on a fixed tabletop. L2 preserves the fixed-table setting but extends the task horizon. The agent must repeatedly locate, pick, and place every object on the table into a basket while tracking progress across multiple manipulation cycles. L3 is the most challenging task and couples navigation with manipulation across two workspaces, requiring the agent to alternate between the two as it travels to the source, identifies and retrieves the instructed object, navigates to the destination, and completes the delivery. Across tasks, we vary object categories and placements. For each method, we conduct 20 independent trials on L1 and L2 and 15 independent trials on L3.

We compare \thea with the following systems: end-to-end policies~\sidecite{zhao2023act,wu2026lingbotvla2,black2025pi05}, a Coding-as-Policy baseline, and a hierarchical planning method based on SayCan~\sidecite{ahn2022saycan}.
The Coding-as-Policy baseline revises its robot-control program over multiple turns using execution traces and structured text describing the initial scene, subsequent visual changes, and task completion.
This follows the CaP-Bench M3 setting in CaP-X~\sidecite{fu2026capx}.
We use GPT-5.5 with reasoning effort set to \code{high} as the decision model for \thea, CaP-X, and SayCan. For perception, \thea builds on the Astribot S1 SDK and SysNav~\sidecite{zhu2026sysnav}.
Each end-to-end policy is trained or fine-tuned on 200 collected demonstrations per task. The Coding-as-Policy and hierarchical planning methods use the same tools and underlying implementations as \thea (Table~\ref{tab:tools}, Appendix~\ref{app:tools}). For the orchestration baselines~\cite{fu2026capx,ahn2022saycan}, this setup controls for the underlying robot capabilities and isolates how each architecture selects and composes tools, and how it recovers when they fail.
Two of the end-to-end baselines, ACT and $\pi_{0.5}$, are the same policies that back \thea's manipulation tools (Table~\ref{tab:tools}).
The differences reported below therefore reflect how complete each architecture's orchestration is, not how capable the underlying policies are.

\subsection{Scaling with Task Complexity}
\label{sec:exp-comparison}

\begingroup
\widefloatsetup
\begin{figure}[!t]
  \centering
  \makebox[\textwidth][l]{%
    \includegraphics[width=\dimexpr\textwidth+\marginparsep+\marginparwidth\relax]{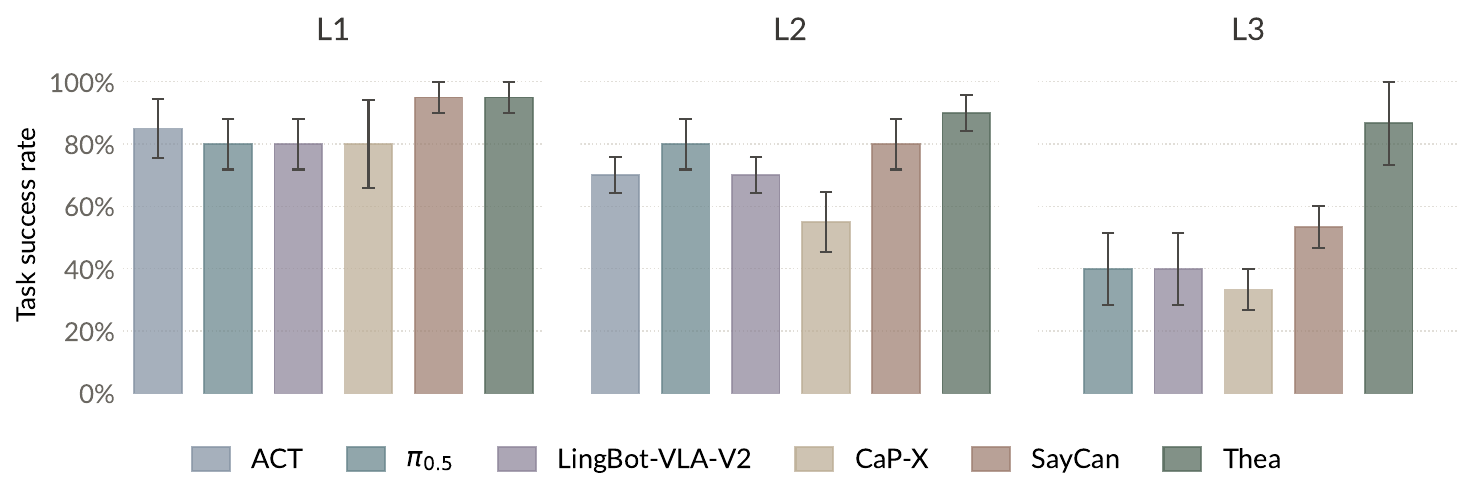}%
  }
  \refstepcounter{figure}
  \label{fig:task-success}
  \smallskip
  \makebox[\textwidth][l]{%
    \parbox{\dimexpr\textwidth+\marginparsep+\marginparwidth\relax}{\raggedright\small\textbf{Figure~\thefigure.} Task success rates of the baselines and \thea across tasks of increasing difficulty. Error bars show one standard error across runs. ACT is omitted from L3 because it is not language-conditioned.\par}%
  }
\end{figure}
\endgroup

Figure~\ref{fig:task-success} reports the task success rates of ACT~\cite{zhao2023act}, LingBot-VLA-V2~\cite{wu2026lingbotvla2}, $\pi_{0.5}$~\cite{black2025pi05}, CaP-X~\cite{fu2026capx}, SayCan~\cite{ahn2022saycan}, and \thea across L1--L3.
Beyond absolute performance, the comparison shows how each architecture scales as task horizon and compositional complexity grow.
A trial counts as successful only if every instruction-specified object reaches its target location and the robot completes all required navigation and manipulation stages; outcomes are judged by human annotators from recorded trajectories, independently of the online evaluator.

\thea achieves the highest success rate at all three levels. On L1 the gap among methods is small. On L2, \thea evaluates each grasp and retries the failed ones instead of carrying an invalid sequence forward. On L3, the agent calls \code{navigate\_to} and then \code{move\_base} to bring the robot to a pose better suited to the next manipulation, and adjusts the pose again when a manipulation fails. For the end-to-end policies (ACT, LingBot-VLA-V2, $\pi_{0.5}$), a failure in the middle of a task goes undetected, and the run does not recover. CaP-X generates and revises programs across multiple turns; SayCan selects the next skill with no verdict on the last. The gap that widens from L1 to L3 follows the reliability arithmetic of \S\ref{sec:arithmetic}, where per-step failures compound with task length.
It also reflects key properties of the harness. The model selects and composes tools as the task requires, and the evaluator closes the loop, making outcomes observable and failures recoverable.

\subsection{Evaluator Accuracy}
\label{sec:exp-evaluator}

The recovery behind the results above rests on the evaluator, as the analysis of \S\ref{sec:arithmetic} suggests.
We therefore measure how reliably the evaluator judges the state of an ongoing physical task.
The test set consists of 90 trajectory checkpoints collected from Astribot S1, AgileX Cobot Magic, and Unitree G1, sampled in equal numbers after successful completion, mid-execution, and after unrecovered failures.
Two human annotators labeled each checkpoint independently from the complete trajectory, resolving disagreements by replay and discussion.
We use Qwen3.7-Plus~\sidecite{qwen37plus} as the evaluator model throughout our experiments.
At test time the evaluator sees only its evaluation prompt, including the per-tool post-condition (the expected world state after the call), and the synchronized camera observations.

\begin{figure}[t]
  \centering
  \includegraphics[width=.76\textwidth]{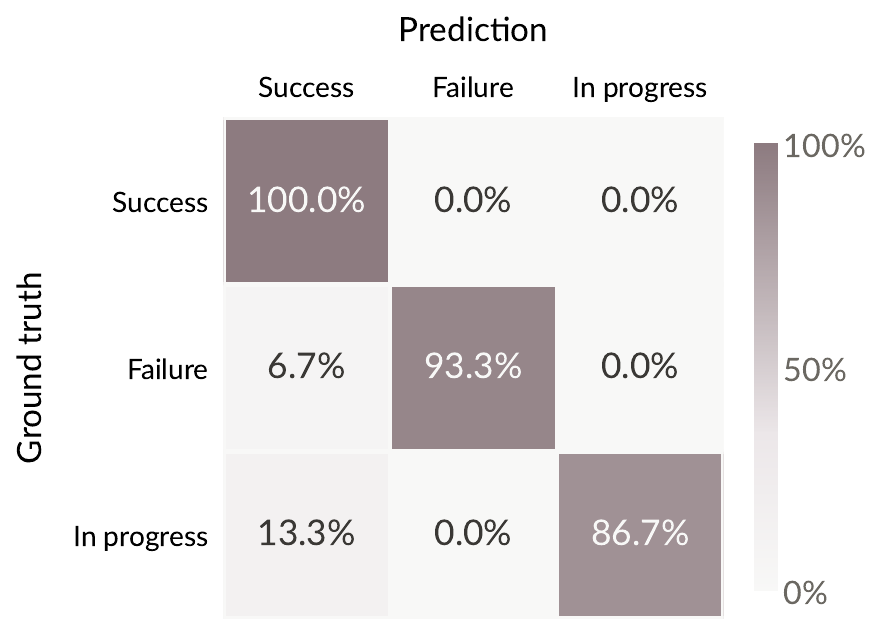}\par
  \refstepcounter{figure}
  \label{fig:evaluator-confusion}
  \smallskip
  {\small\raggedright\noindent\textbf{Figure~\thefigure.} Evaluator predictions against ground-truth success, failure, and in-progress states across three robot embodiments.\par}
\end{figure}

Figure~\ref{fig:evaluator-confusion} reports the classification performance for each state separately, with an average accuracy of 93.3\%.
The remaining errors all run in one direction: 6.7\% of failed and 13.3\% of in-progress checkpoints are judged successful, while no successful checkpoint is misjudged.
These errors are the harmful kind, because a false success either forfeits a needed retry or cuts an action short.
Reducing this error mode is an important next step.
Substituting the measured accuracy ($\alpha = 0.93$)\sidenote{The three-state average, taken as an approximation of the binary $\alpha$ defined in \S\ref{sec:arithmetic}.}, per-step success $p = 0.8$, and up to $k = 5$ retries into the formulation of \S\ref{sec:arithmetic}, a five-step task completes with probability $0.33$ in open loop and $0.91$ under the harness, broadly consistent with the trend in Figure~\ref{fig:task-success}.

\subsection{Emergent Capabilities}
\label{sec:exp-capabilities}

This subsection moves from the fixed tasks above to longer, open-ended tasks in more complex, realistic settings.
\thea runs with the full harness, every tool and component available, and we observe what the agent as a whole composes out of them.
The demonstrations below sample the capabilities that emerge, and Appendix~\ref{app:demo-logs} provides the execution logs of the demos.

\paragraph{Long-horizon task composition.}
Across these deployments, \thea carries user requests through to completion over extended sequences of physical interaction.
The difficulty concentrates where navigation and manipulation alternate, because each manipulation depends on where the previous move left the robot.
\thea composes the alternation from its tools (Figure~\ref{fig:emergent-capabilities}\subref{sub:emergent-longhorizon}): it reads the target's position from the scene graph, approaches with \code{navigate\_to}, refines the base pose with \code{move\_base} within the measured clearances, and hands each manipulation's outcome to the evaluator, whose verdict decides whether to adjust and retry or move on.
Figure~\ref{fig:emergent-capabilities} shows representative trajectories, each composed at run time from the same general tools; none follows a script prepared for the task.

\begin{figure}[!t]
  \RawFloats
  \centering
  \refstepcounter{figure}
  \label{fig:emergent-capabilities}
  \setcounter{subfigure}{0}
  \captionsetup[subfigure]{
    format=plain,
    font=small,
    labelfont=bf,
    labelformat=parens,
    labelsep=period,
    justification=raggedright,
    singlelinecheck=no,
    skip=3pt,
  }

  \begin{subfigure}[t]{\textwidth}
    \centering
    \includegraphics[width=.318\linewidth]{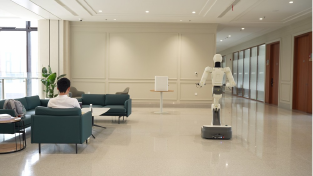}\hfill
    \includegraphics[width=.318\linewidth]{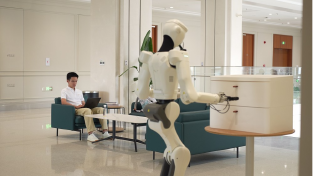}\hfill
    \includegraphics[width=.318\linewidth]{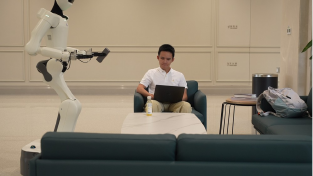}
    \caption{The robot executes a long-horizon sequence of navigation and manipulation, composed at run time, to retrieve and deliver a power bank.}
    \label{sub:emergent-longhorizon}
  \end{subfigure}

  \vspace{5pt}
  \begin{subfigure}[t]{\textwidth}
    \centering
    \includegraphics[width=.318\linewidth]{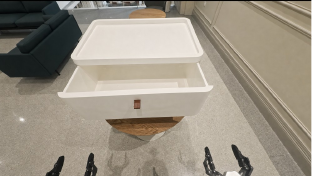}\hfill
    \includegraphics[width=.318\linewidth]{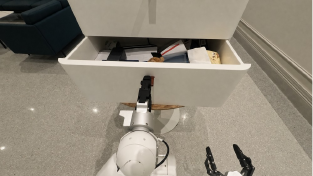}\hfill
    \includegraphics[width=.318\linewidth]{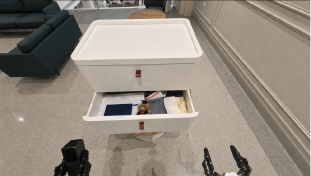}
    \caption{The robot actively searches the cabinet drawers to locate the power bank when the target is absent from the scene graph.}
    \label{sub:emergent-active}
  \end{subfigure}

  \vspace{5pt}
  \begin{subfigure}[t]{\textwidth}
    \centering
    \includegraphics[width=.318\linewidth]{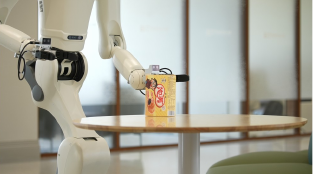}\hfill
    \includegraphics[width=.318\linewidth]{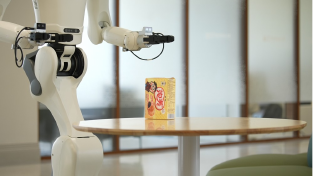}\hfill
    \includegraphics[width=.318\linewidth]{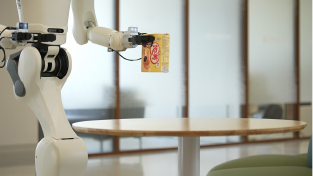}
    \caption{The robot uses the evaluator's failure reason to reposition relative to the object and complete the grasp after the initial attempt fails.}
    \label{sub:emergent-recovery}
  \end{subfigure}

  \vspace{5pt}
  \begin{subfigure}[t]{\textwidth}
    \centering
    \includegraphics[width=.318\linewidth]{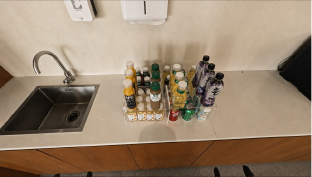}\hfill
    \includegraphics[width=.318\linewidth]{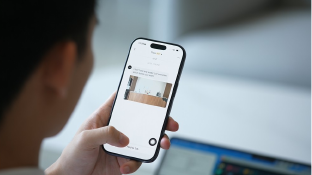}\hfill
    \includegraphics[width=.318\linewidth]{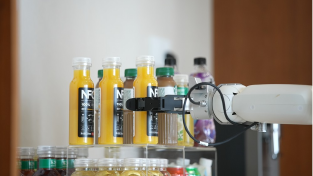}
    \caption{The robot queries the user for an alternative when the requested water is unavailable, then retrieves the selected drink.}
    \label{sub:emergent-user}
  \end{subfigure}

  \vspace{5pt}
  \begin{subfigure}[t]{\textwidth}
    \centering
    \includegraphics[width=.318\linewidth]{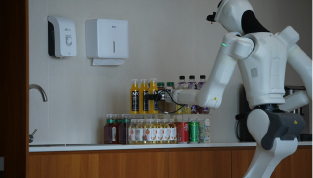}\hfill
    \includegraphics[width=.318\linewidth]{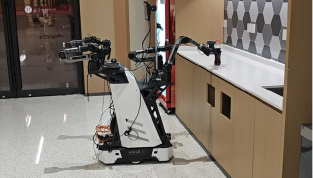}\hfill
    \includegraphics[width=.318\linewidth]{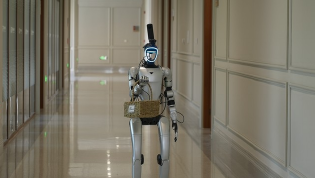}
    \caption{The three robots share the same harness, with embodiment-specific profiles and tool implementations.}
    \label{sub:emergent-portability}
  \end{subfigure}

  \smallskip
  \addtocounter{figure}{-1}
  {\small\raggedright\noindent\textbf{Figure~\thefigure.} Demonstrations of emergent capabilities enabled by the shared harness across long-horizon task composition, active perception, failure recovery, user interaction, and cross-embodiment portability.\par}
\end{figure}

\paragraph{Active perception.}
Active perception takes two forms.
When a requested object appears nowhere in the scene graph, \thea explores the environment. It navigates to a cabinet, opens and inspects its drawers in sequence (Figure~\ref{fig:emergent-capabilities}\subref{sub:emergent-active}), and writes the previously occluded contents into the graph.
When the evidence lies outside the current view, \thea adjusts the view instead, calling \code{tilt\_head} to bring the relevant region into the camera.
The two operate at different scales, moving the body and moving the camera, but follow the same logic: perception is an action the model chooses, and the new perception informs the next decision.

\paragraph{Failure recovery.}
Physical execution does not always satisfy the intended post-condition on the first attempt.
When it does not, the evaluator returns the failure with a reason, and \thea decides the next call from that reason and the surrounding evidence (Figure~\ref{fig:emergent-capabilities}\subref{sub:emergent-recovery}), perhaps adjusting the base pose before retrying, or switching to a tool backed by a different policy.
When a retry succeeds, the consolidation at task end writes the recovery into that tool's experience.

\paragraph{User interaction.}
\thea keeps the user reachable throughout the task.
A request may prove vague, a choice may hinge on the user's preference, or progress may be worth surfacing; \code{query\_user} and \code{notify\_user} serve these moments, and the model calls them when the need arises.
In the beverage scene (Figure~\ref{fig:emergent-capabilities}\subref{sub:emergent-user}), the user asks for a bottle of water and the scene graph holds none; rather than grasping a substitute of its own choosing, \thea asks the user which of the drinks it does see to bring, and retrieves the one the user selects.

\paragraph{Cross-embodiment portability.}
The same harness runs on all three embodiments, Astribot S1, AgileX Cobot Magic, and Unitree G1 (Figure~\ref{fig:emergent-capabilities}\subref{sub:emergent-portability}).
Nothing in the loop refers to a particular body; what is body-specific enters through the Embodiment Profile and the tool implementations behind the registry.
Moving to a new robot therefore means writing its profile and its tools, while the loop above them, interpreting requests, selecting tools, weighing evaluator verdicts, carries over unchanged.

\FloatBarrier

\section{Related Work}
\label{sec:related}

\subsection{Coding Agents}
\label{sec:rw-coding}

Modern coding agents succeed at managing an environment end to end at scale~\sidecite{anthropic2026claude,openai2026codex,wang2024openhands}.
Their shared design has been systematized as a harness~\cite{weng2026harness}: a reactive loop that re-decides from the latest observed state~\sidecite{yao2023react}, a tool registry the model selects from at runtime~\sidecite{schick2023toolformer}, recovery that reads a failure trace and retries~\sidecite{shinn2023reflexion}, managed context~\sidecite{packer2023memgpt}, and memory that persists across sessions~\sidecite{wang2023voyager}.
We transplant these patterns into the physical world, adapting each component along the way (\S\ref{sec:system}).
Two properties of the software environment have no physical counterpart.
A code repository is readable, and its tests are verifiable by construction.
The physical world provides neither, so we build both (\S\ref{sec:gaps}).

\subsection{Embodied Foundation Models}
\label{sec:rw-efm}

Embodied foundation models supply atomic capability in the physical world along two complementary lines.
Vision-Language-Action (VLA) models map observation and instruction end to end to actions, from the RT series~\sidecite{zitkovich2023rt2} through open generalist policies~\sidecite{kim2024openvla} to flow-matching architectures~\cite{black2025pi0,black2025pi05}.
World Action Models (WAMs) grow out of world models, predictors of how the world evolves, and fold action emission into the prediction network. DreamZero, a recent instance, drives a robot in closed loop from a single image-to-video diffusion model~\sidecite{ye2026dreamzero}.
Both lines are the tools our harness calls, not the layer it replaces. A recent evaluation across control interfaces finds that frontier models mostly fail when driving joints directly but act capably when supervising pretrained policies, with the interface mattering as much as the model itself~\cite{berman2026robotics}.
Today's best policies succeed on a single step with probability $p$ of roughly $0.8$ to $0.9$~\cite{kim2024openvla,black2025pi05}, so an $n$-step task succeeds with only $p^n$ (\S\ref{sec:arithmetic}), which collapses even for modest $n$; recovery and evaluation at the harness level close this gap without waiting for single-step perfection.
This is why our design separates atomic capability from orchestration explicitly.

\subsection{Orchestration for Embodied Agents}
\label{sec:rw-orch}

Language-model orchestration over robot skills begins open-loop. SayCan selects affordance-grounded skills step by step but nothing judges the skill just executed~\sidecite{ahn2022saycan}, and Code as Policies emits a one-shot program~\sidecite{liang2023code}.
Inner Monologue comes closest to closing the loop, threading environment feedback back through the language model, but the feedback channel is hand-picked and there is no general mechanism to decide whether a step has actually succeeded~\sidecite{huang2022inner}.
Later work supplies individual pieces of the loop, a scene graph the planner reads~\sidecite{rana2023sayplan} and a failure judge that rules on outcomes~\sidecite{duan2025aha}.
Dual-system VLAs take a different route and pair a slow vision-language reasoner with a fast low-level controller~\sidecite{nvidia2025groot,figure2025helix,shi2025hirobot,gemini2025robotics}.
Most recently, concurrent systems bring coding agents to robotics: RoboClaw automates data collection through self-resetting action pairs~\sidecite{li2026roboclaw}, CaP-X benchmarks coding agents on manipulation and improves them by scaling test-time interaction~\sidecite{fu2026capx}, Guava searches the harness design space for manipulation and distills the result into a compact model for deployment~\sidecite{liu2026guava}, ENPIRE has coding agents self-improve policies on real robots through automated reset, rollout, and verification~\sidecite{nvidia2026enpire}, and ASPIRE discovers reusable skills by writing and repairing control code~\sidecite{lu2026aspire}.
Our work carries the paradigm of coding agents over as a whole. The harness is itself the deployed system, closing the loop from instruction to completion in the physical world.

\section{Discussion}
\label{sec:discussion}

This work took an exploratory step towards the harness of embodied agents, realizing the paradigm of coding agents as one working system, \thea.
Each component was adapted to the physical world, from the loop and tools to context and memory. The two properties the world does not grant, readability and verifiability, were rebuilt as the scene graph and the evaluator.

We offer \thea as evidence that embodied AI is entering the paradigm shift that software agents have already undergone, where the layer between the model and its environment has come to matter as much as the model itself~\sidecite{weng2026harness,lopopolo2026harness,yang2024sweagent}.
In fact, an embodied agent needs such a system, rather than just a capable model, even more than a coding agent does, as what it faces is far more complex.
The world changes and surprises on its own schedule, observations are always partial, executions fail and cannot be rolled back, and the body's limits shape what is possible.

We believe this shift will fundamentally change how embodied agents are built:
\begin{itemize}
  \item Reliability becomes a systems problem.
  It can be engineered in the harness without touching model weights, and the gain compounds with the model's own progress, since a stronger model raises what the same harness delivers.
  \item Capability becomes additive.
  A new policy enters as one more callable tool, and a new body as one more embodiment profile, with nothing retrained around them, so policy builders and system builders can advance in parallel.
  \item The loop becomes a training target.
  Reading a scene graph, choosing among tools, admitting failure and retrying are today the work of a general language model. They are learnable behaviors, and natural objectives for the next generation of embodied foundation models. VLAs and WAMs enter the loop as tools today, and they can evolve towards the objectives the loop sets.
\end{itemize}

Necessary as we have argued each component of the harness to be, we do not claim any of the specific implementations to be final.
Indeed, each has its own limitations.
The scene graph is limited by the perception beneath it.
Positions are coarse, and associating observations of the same object over time still admits errors.
The evaluator approximates the exit code it replaces, and its verdicts carry false positives and false negatives that a return value does not.
And the loop takes time to decide.
Every decision passes through a language model, and the system does not yet act in real time.

Each of these limits marks a direction the framework can grow.
For the scene graph, we can explore representations of the world that are more efficient and more durable, tracking entities through change, towards a household memory of where things are and who moved them.
The evaluator will always carry an error rate, but the error rate can be trained down, with large collections of judged outcomes, with the habit of gathering more evidence before ruling, and in the long run jointly with the policies it judges.
A compact evaluation model, trained for this one job and run locally, could then rule on every action without a round trip to a frontier foundation model.
A robot, more than a coding agent, has reason to run its models on board, so a compact model trained for the loop, accelerated at the edge, would bring decisions towards real time.
Memory also changes what the agent is to the people it serves.
It can grow more personalized and more proactive over time, learning a household's habits and offering help before being asked.
Such familiarity raises new questions of privacy, trust, and initiative that deserve as much care as capability.
And deployment itself feeds improvement.
Interaction with the real world produces exactly what policy learning consumes, ready for the agentic self-improvement engines built by recent work~\sidecite{nvidia2026enpire,lu2026aspire}, closing a flywheel from experience to policies to stronger single steps.
We envision \thea growing with each turn of this flywheel, into part of the infrastructure that embodied agents take for granted.

\clearpage
\printbibliography

\clearpage
\appendix
\makeatletter\setlength{\@fptop}{0pt}\makeatother

\section{Tool Registry}
\label{app:tools}
Table~\ref{tab:tools} lists the Astribot S1 tool registry.
Its entries follow the protocol of \S\ref{sec:tool-protocol}: physical capabilities use deployment backends, while lightweight context and interaction tools run in-process.
Slash-separated lists are the admissible values; quoted strings are free-form descriptions; refs like \code{desk\_24} name Scene Graph entities.

\begingroup
\small
\newlength{\toolregistrywidth}
\setlength{\toolregistrywidth}{\dimexpr\linewidth+\marginparsep+\marginparwidth\relax}
\setlength{\linewidth}{\toolregistrywidth}
\setlength{\hsize}{\toolregistrywidth}
\setlength{\LTleft}{0pt}
\setlength{\LTright}{0pt}
\setlength{\LTpre}{6pt}
\setlength{\LTpost}{6pt}
\renewcommand{\arraystretch}{1.08}
\setlength{\LTcapwidth}{\toolregistrywidth}
\newcommand{\toolnote}[1]{{\footnotesize\itshape #1}}
\par\medskip
\noindent\parbox{\toolregistrywidth}{\raggedright\small\textbf{Table~\ref{tab:tools}.} Tools deployed on the Astribot S1, grouped by function. \code{evaluate\_run} is registered alongside them but invoked by the harness through a post-hook, never selected by the model (\S\ref{sec:eval}).\par}
\par\smallskip
\begin{longtable}{@{}
  >{\raggedright\arraybackslash}p{0.145\toolregistrywidth}
  @{\hspace{2pt}}
  >{\raggedright\arraybackslash}p{0.22\toolregistrywidth}
  @{\hspace{2pt}}
  >{\raggedright\arraybackslash}p{0.18\toolregistrywidth}
  @{\hspace{2pt}}
  >{\raggedright\arraybackslash}p{0.20\toolregistrywidth}
  @{\hspace{2pt}}
  >{\raggedright\arraybackslash}p{\dimexpr0.255\toolregistrywidth-8pt\relax}
  @{}}
  \noalign{\label{tab:tools}}
  \toprule
  Tool & Description & Backend & Parameter & Value \\
  \midrule
  \endfirsthead
  \caption*{Table~\ref{tab:tools}. Tools deployed on the Astribot S1 (continued).}\\
  \toprule
  Tool & Description & Backend & Parameter & Value \\
  \midrule
  \endhead
  \midrule
  \multicolumn{5}{r@{}}{\textit{Continued on next page}}\\
  \endfoot
  \bottomrule
  \endlastfoot

  \multicolumn{5}{@{}l}{\textbf{Navigation}} \\
  \code{navigate\_to} &
  Approach an object or location. &
  SysNav~\cite{zhu2026sysnav} &
  \code{target} &
  \eg ``drink area''/desk\_24 \\
  \code{move\_base} &
  Translate or rotate within safety bounds. &
  Astribot S1 SDK &
  \code{direction} &
  forward/\allowbreak backward/\allowbreak
  left/\allowbreak right/\allowbreak
  turn\_left/\allowbreak turn\_right \\*
  {} & {} & {} &
  \code{distance\_m} &
  0.03--0.80\,m \\*
  {} & {} & {} &
  \code{angle\_deg} &
  3--30$^\circ$ \\
  \midrule

  \multicolumn{5}{@{}l}{\textbf{Manipulation}} \\
  \code{pick\_up} &
  Grasp a localized object. &
  CaP-X/ACT/$\pi_{0.5}$\newline
\cite{fu2026capx,zhao2023act,black2025pi05} &
  \code{obj} &
  \eg ``orange juice''/bottle\_2 \\
  \code{place} &
  Place the held object on a table or in a box. &
  CaP-X/ACT/$\pi_{0.5}$ &
  \code{target} &
  \eg ``table''/table\_2 \\
  \code{open\_drawer} &
  Open a drawer. &
  CaP-X/ACT/$\pi_{0.5}$&
  \code{container\_ref} &
  \eg cabinet\_87 \\*
  {} & {} & {} &
  \code{drawer\_level} &
  higher/\allowbreak lower \\
  \code{close\_drawer} &
  Close a drawer. &
  CaP-X/ACT/$\pi_{0.5}$ &
  \code{container\_ref} &
  \eg cabinet\_87 \\*
  {} & {} & {} &
  \code{drawer\_level} &
  higher/\allowbreak lower \\
  \code{trash\_drop} &
  Drop the held object into a bin. &
  CaP-X/ACT/$\pi_{0.5}$ &
  \code{trash\_ref} &
  \eg trash\_can\_54 \\
  \midrule

  \multicolumn{5}{@{}l}{\textbf{Perception}} \\
  \code{tilt\_head} &
  Tilt the head camera to a given pitch angle. &
  Astribot S1 SDK &
  \code{pitch\_deg} &
  0--60$^\circ$ \\
  \code{get\_object\_}\newline \code{relations} &
  Query an object's spatial and semantic relations. &
  Scene Graph &
  \code{obj}\newline
  \code{relation}~\toolnote{(optional)} &
  \eg cup\_2\newline
  on/\allowbreak inside/\allowbreak
  holding/\allowbreak near \\
  \code{get\_image} &
  Retrieve stored images of an object. &
  Scene Graph &
  \code{obj} &
  \eg cup\_2 \\
  \midrule

  \multicolumn{5}{@{}l}{\textbf{Skills}} \\
  \code{load\_skill} &
  Load the instructions of a registered skill. &
  \thea &
  \code{name} &
  \eg tidy-workspace \\
  \midrule

  \multicolumn{5}{@{}l}{\textbf{User Interaction}} \\
  \code{query\_user} &
  Request and await clarification. &
  Lark &
  \code{question} &
  \eg ``No plain water; which drink should I bring?'' \\*
  {} & {} & {} &
  \code{candidate\_refs}\newline
  \toolnote{(optional)} &
  \eg bottle\_3 \\*
  {} & {} & {} &
  \code{observation\_views}\newline
  \toolnote{(optional)} &
  \eg torso\_rgbd \\
  \code{notify\_user} &
  Send a non-blocking update. &
  Lark &
  \code{message} &
  \eg ``I found an empty orange-juice bottle on desk...'' \\*
  {} & {} & {} &
  \code{notification\_type}\newline
  \toolnote{(optional)} &
  progress/\allowbreak warning/\allowbreak
  completion \\
\end{longtable}
\endgroup

\section{Context Structure}
\label{app:context}
Section~\ref{sec:context} introduces three context lifetimes.
Listing~\ref{lst:context-structure} expands the context items defined in
Table~\ref{tab:context-lifetimes}. The System Prompt appears in
full; other entries show structure without task-specific values.

\newcommand{\contextkey}[1]{\textcolor{codebuiltin}{#1}}
\newcommand{\contextoptional}[1]{\textcolor{black!46}{#1}}
\newcommand{\contextentry}[2]{%
  {\rmfamily\normalsize\color{contextink}#1}\par\vspace{.18em}
  {\raggedright #2\par}%
}
\newcommand{\contextgap}{\vspace{1.1em}}
\newcommand{\contextpromptbullet}[1]{%
  \par\hangindent=1.1em\hangafter=1\noindent - #1\par\vspace{.08em}%
}

\newtcolorbox{contextpanel}[4]{%
  enhanced, boxrule=0pt, frame hidden, sharp corners,
  colback=#1,
  left=9pt, right=9pt, top=6pt, bottom=7pt,
  before skip=0pt, after skip=5pt,
  overlay={\node[anchor=north west, inner sep=0pt,
      xshift=\marginparsep, text width=\marginparwidth, align=left]
    at (frame.north east)
    {\footnotesize{\bfseries\textcolor{#2}{#3}}\\\textcolor{#2}{#4}};},
}

\refstepcounter{lstlisting}\label{lst:context-structure}%
\medskip

\begin{contextpanel}{contextresidentbg}{contextresidentink}{Resident}{stable across turns}
\ttfamily\small
\colorlet{contextink}{contextresidentink}%
\contextentry{System Prompt}{%
You are Thea, an embodied agent that completes physical tasks by calling
tools.\par\vspace{.35em}
\contextpromptbullet{Choose one next tool call per turn unless the task
is complete.}
\contextpromptbullet{After each tool result, update the plan from the
latest evidence.}
\contextpromptbullet{Use the Scene Graph Brief for object refs and coarse
global state. Use the latest Observation for current local visibility,
alignment, and clearance.}
\contextpromptbullet{Treat the latest Observation as current physical
evidence and earlier tool results as historical execution evidence.}
\contextpromptbullet{Follow tool descriptions for preconditions,
argument semantics, failure modes, and recovery hints.}
\contextpromptbullet{Do not invent object identifiers, measurements,
action outcomes, or user intent. Ask the user when the instruction and
current evidence are genuinely insufficient.}
\contextpromptbullet{Wait for the current tool to finish before choosing
the next tool.}}
\contextgap
\contextentry{Memory}{%
\{\contextkey{Preferences}: [entry, \ldots],\quad
\contextkey{Conventions}: [entry, \ldots],\\
\hspace*{1em}\contextkey{General Lessons}: [entry, \ldots]\}}
\contextgap
\contextentry{Embodiment Profile}{%
\contextkey{Operational Envelope}:
\{Base Footprint; Base Mobility; Reachable Workspace\}\\
\contextkey{Perception Configuration}:
\{Sensor Modalities; Model-Visible Views\}\\
\contextkey{Base-Relative Positions}:
\{Camera Positions; Initial Gripper Positions\}}
\contextgap
\contextentry{Tool Definitions}{%
[\,\{\contextkey{name},
\contextkey{description}
\contextoptional{[+ experience summary]},\\
\hspace*{1em}\contextkey{inputSchema}:
\{\contextkey{type}, \contextkey{properties},
\contextoptional{[\contextkey{required}]}\}\}, \ldots\,]}
\end{contextpanel}

\begin{contextpanel}{contextrefreshedbg}{contextrefreshedink}{Refreshed}{replaced before each decision}
\ttfamily\small
\colorlet{contextink}{contextrefreshedink}%
\contextentry{Scene Graph Brief}{%
\contextkey{Graph metadata}:
freshness=\ldots; provenance=\ldots;
coordinate\_frame=\ldots; updated\_at=\ldots\\
\contextkey{Robot}: pose=(\ldots); holding=[obj, \ldots]\\
\contextkey{Objects}: - obj; position=(\ldots); confidence=\ldots;
freshness=\ldots; container\_state=\ldots; contents=[obj, \ldots]}
\contextgap
\contextentry{Observations}{%
\contextkey{Base clearance}:
forward=\ldots\,m, backward=\ldots\,m,
left=\ldots\,m, right=\ldots\,m\\
\contextkey{Visual evidence}: view, \ldots\quad
\contextoptional{[image blocks follow]}}
\end{contextpanel}

\begin{contextpanel}{contextaccumulatedbg}{contextaccumulatedink}{Accumulated}{grows through ordinary turns}
\ttfamily\small
\colorlet{contextink}{contextaccumulatedink}%
\contextentry{Instructions}{%
[\,\{\contextkey{content}: instruction\}, \ldots\,]}
\contextgap
\contextentry{Task Notes}{%
\contextkey{Task summary}:
\contextkey{Goal}=instruction;
\contextkey{Current phase}=phase\\
\contextkey{Timeline}: task\_start; tool: outcome
\contextoptional{[reason]}}
\contextgap
\contextentry{Model Responses}{%
[\,\{\contextoptional{\contextkey{content}},\\
\hspace*{1em}\contextoptional{\contextkey{tool\_calls}[]}:
\{\contextkey{id}, \contextkey{type}: function,
\contextkey{function}:
\{\contextkey{name}, \contextkey{arguments}\}\}\\
\}, \ldots\,]}
\contextgap
\contextentry{Tool Results}{%
[\,\{\contextkey{tool\_call\_id},\\
\hspace*{1em}\contextkey{content}:
\{\contextkey{success}: true,
\contextoptional{tool-defined value fields}\}
\contextoptional{or }
\{\contextkey{success}: false, \contextkey{reason}\}\\
\}, \ldots\,]}
\contextgap
\contextoptional{Compaction may replace an older accumulated prefix;
resident and refreshed context remain unchanged.}\par
\end{contextpanel}

\par\nopagebreak\smallskip\nopagebreak
{\small\textbf{Listing~\thelstlisting.} Context structure for one
decision, the concatenation of the three lifetimes of
Table~\ref{tab:context-lifetimes} in one model call.}

\section{Demo Execution Logs}
\label{app:demo-logs}
The first three logs were collected on the Astribot S1, and the final log on
the Unitree G1 with the same harness.
Turn counts follow the definition in \S\ref{sec:agent-loop}. Final responses
contain no tool calls and are therefore shown without turn numbers.

\subsection{Beverage Retrieval}
\label{app:beverage-log}

The user asks for a bottle of water to be brought to a table. The beverage
area contains no unique water bottle, and \thea asks the user to select an
available drink. After the user chooses orange juice, the robot delivers it.

\begingroup

\definecolor{traceStageBg}{HTML}{E7E8E5}
\definecolor{traceStageInk}{HTML}{4A4B47}
\definecolor{traceMuted}{HTML}{6B6862}
\definecolor{traceRule}{HTML}{C9C6BF}
\definecolor{traceHookBg}{HTML}{EDF1EB}
\definecolor{traceHookInk}{HTML}{55784D}
\definecolor{traceNavBg}{HTML}{E7ECF6}
\definecolor{traceNavInk}{HTML}{4D5C78}
\definecolor{traceManipBg}{HTML}{F3F0EA}
\definecolor{traceManipInk}{HTML}{786A4D}
\definecolor{traceEvaluatorBg}{HTML}{F5E8EC}
\definecolor{traceEvaluatorInk}{HTML}{784D59}
\definecolor{traceResponseBg}{HTML}{EBE8F5}
\definecolor{traceResponseInk}{HTML}{564D78}
\definecolor{traceUserBg}{HTML}{EBE8F5}
\definecolor{traceUserInk}{HTML}{564D78}
\definecolor{traceFailureBg}{HTML}{F0E3E1}
\definecolor{traceFailureInk}{HTML}{6E4340}

\lstdefinestyle{appendixtrace}{
  style=thea,
  aboveskip=.45em,
  belowskip=.55em,
  framextopmargin=3pt,
  framexbottommargin=3pt,
}
\lstset{style=appendixtrace}

\lstnewenvironment{tracecode}[1][]
  {\lstset{firstnumber=1,aboveskip=0pt,belowskip=0pt,#1}%
   \endgraf\addvspace{.45em}\noindent\minipage{\linewidth}}
  {\endminipage\endgraf\addvspace{.55em}}

\newcommand{\appendixtracereasoning}[2]{%
  \noindent{\small\textbf{Turn #1.}\enspace
  \textbf{Reasoning:} #2}\par\nobreak
}
\newcommand{\appendixtracegroupreasoning}[2]{%
  \noindent{\small\textbf{Turns #1.}\enspace
  \textbf{Reasoning summary:} #2}\par\nobreak
}
\newcommand{\appendixtracegroupreasoningfirst}[3]{%
  \noindent{\small\textbf{Turns #1.}\enspace
  \textbf{Reasoning summary:}\sidenote{#3} #2}\par\nobreak
}
\newcommand{\appendixtracefoldin}[4]{%
  \vspace{.3em}%
  {\setlength{\fboxsep}{6pt}%
    \noindent\makebox[\linewidth][c]{%
      \colorbox{#1}{%
        \parbox{\dimexpr\linewidth+16pt-2\fboxsep\relax}{%
          {\small
            \textbf{\textcolor{#2}{#3}}\hfill
            \textcolor{traceMuted}{#4}}%
        }%
      }%
    }\par}\nobreak\vspace{.4em}%
}
\newcommand{\appendixtracefold}[2]{\appendixtracefoldin{traceNavBg}{traceNavInk}{#1}{#2}}
\newcommand{\appendixtracenote}[2]{%
  {\setlength{\fboxsep}{6pt}%
    \noindent\makebox[\linewidth][c]{%
      \colorbox{#1}{%
        \parbox{\dimexpr\linewidth+16pt-2\fboxsep\relax}{%
          {\small #2}%
        }%
      }%
    }\par}\vspace{.5em}%
}
\newcommand{\appendixtracehooktoken}[1]{%
  {\ttfamily\textcolor{traceHookInk}{#1}}%
}
\newcommand{\appendixtraceevaluatortoken}[1]{%
  {\ttfamily\textcolor{traceEvaluatorInk}{#1}}%
}
\newcommand{\appendixtraceautocalls}[1]{%
  \appendixtracenote{traceHookBg}{%
    \textbf{\textcolor{traceHookInk}{Automatic calls.}} #1}%
}
\newcommand{\appendixtracenavresult}[2]{%
  \appendixtracenote{traceNavBg}{%
    \textbf{\textcolor{traceNavInk}{#1}} #2}%
}
\newcommand{\appendixtracemanipresult}[2]{%
  \appendixtracenote{traceManipBg}{%
    \textbf{\textcolor{traceManipInk}{#1}} #2}%
}
\newcommand{\appendixtracecommresult}[2]{%
  \appendixtracenote{traceUserBg}{%
    \textbf{\textcolor{traceUserInk}{#1}} #2}%
}
\newcommand{\appendixtraceevaluatorresult}[2]{%
  \appendixtracenote{traceEvaluatorBg}{%
    \textbf{\textcolor{traceEvaluatorInk}{Evaluator.}}
    The harness invokes
    \appendixtraceevaluatortoken{evaluate\_run()} after the manipulation.\\[-.05em]
    \textbf{\textcolor{traceEvaluatorInk}{#1}} #2}%
}
\newcommand{\appendixtraceactionevaluatorresult}[2]{%
  \appendixtracenote{traceEvaluatorBg}{%
    \textbf{\textcolor{traceEvaluatorInk}{Evaluator.}}
    The harness invokes
    \appendixtraceevaluatortoken{evaluate\_run()} after the action.\\[-.05em]
    \textbf{\textcolor{traceEvaluatorInk}{#1}} #2}%
}
\newcommand{\appendixtracefailureresult}[2]{%
  \appendixtracenote{traceFailureBg}{%
    \textbf{\textcolor{traceFailureInk}{#1}} #2}%
}
\newcommand{\appendixtracefinal}[1]{%
  {\setlength{\fboxsep}{6pt}%
    \noindent\makebox[\linewidth][c]{%
      \colorbox{traceResponseBg}{%
        \parbox{\dimexpr\linewidth+16pt-2\fboxsep\relax}{%
          {\small\textbf{\textcolor{traceResponseInk}{Final response.}} #1}%
        }%
      }%
    }\par}
}
\newcommand{\appendixtracerequest}[1]{%
  \appendixtracenote{traceUserBg}{%
    \textbf{\textcolor{traceUserInk}{User request.}} #1}%
}

\medskip
\noindent{\color{traceRule}\rule{\linewidth}{1.2pt}}\par
\smallskip
\appendixtracerequest{``Bring me a bottle of water and place it on the table.''}

\begin{samepage}
\appendixtracereasoning{1}{The user requests a drink, so the pantry is the
next search location. Navigate there before selecting the target bottle.}
\begin{tracecode}
navigate_to(target="pantry")
\end{tracecode}
\appendixtracenavresult{Navigation.}{%
The robot reaches the pantry.}
\end{samepage}
\smallskip

\begin{samepage}
\appendixtracereasoning{2}{The navigation result confirms arrival and
alignment at the pantry, and the refreshed clearance evidence permits a
0.80\,m forward step. Advance to improve the torso-camera view and
manipulation distance.}
\begin{tracecode}
move_base(direction="forward", distance_m=0.80)
\end{tracecode}
\end{samepage}
\smallskip

\begin{samepage}
\appendixtracereasoning{3}{The fresh torso view shows no unambiguous plain
water bottle but includes several alternative drinks. Query the user rather
than treating the navigation anchor as the grasp target.}
\begin{tracecode}
query_user(
    question="No plain water; which drink should I bring?")
\end{tracecode}
\appendixtracenote{traceUserBg}{%
\textbf{\textcolor{traceUserInk}{User response.}} Orange juice.}
\end{samepage}
\smallskip

\begin{samepage}
\appendixtracegroupreasoningfirst{4--6}{The user selects orange juice. Apply two
leftward corrections followed by a 0.28\,m forward step to bring the selected
bottle into the grasp envelope.}{Each \code{move\_base} is one turn, decided after the previous completes. Folded runs show the move count and one reasoning summary, here and in the logs that follow.}
\appendixtracefold{Base motions}{%
3 moves}
\begin{tracecode}
move_base(direction="left", distance_m=0.30)
move_base(direction="left", distance_m=0.15)
move_base(direction="forward", distance_m=0.28)
\end{tracecode}
\end{samepage}
\smallskip

\begin{samepage}
\appendixtracereasoning{7}{The user response and local alignment establish
orange juice as the target, but no reliable object reference is available.
Bind it by name, execute the grasp, and verify control.}
\begin{tracecode}
pick_up(obj="orange juice")
\end{tracecode}
\appendixtraceevaluatorresult{Grasp verified.}{%
The evaluator marks the pick as successful.}%
\marginnote{%
  \footnotesize
  The manipulation calls in these logs run self-terminating Coding-as-Policy backends, so their evaluations take the two terminal states of the contract in \S\ref{sec:eval}.%
}
\end{samepage}
\smallskip

\begin{samepage}
\appendixtracereasoning{8}{The grasp evaluator confirms control of the orange
juice. The destination is \code{desk\_24}; navigate there while retaining the
object.}
\begin{tracecode}
navigate_to(target="desk_24")
\end{tracecode}
\appendixtracenavresult{Navigation.}{%
The robot reaches \code{desk\_24} with the drink.}
\end{samepage}
\smallskip

\begin{samepage}
\appendixtracegroupreasoning{9--12}{The navigation result confirms arrival at
\code{desk\_24} with the drink held. Apply a 0.58\,m approach followed by
0.05\,m, 0.04\,m, and 0.03\,m corrections to reach the placement pose.}
\appendixtracefold{Base motions}{%
4 moves}
\begin{tracecode}
move_base(direction="forward", distance_m=0.58)
move_base(direction="forward", distance_m=0.05)
move_base(direction="forward", distance_m=0.04)
...
\end{tracecode}
\end{samepage}
\smallskip

\begin{samepage}
\appendixtracereasoning{13}{The orange juice remains controlled, the final
base correction has completed, and the destination is \code{desk\_24}. Place
the drink there and verify the resulting object state.}
\begin{tracecode}
place(target="table")
\end{tracecode}
\appendixtraceevaluatorresult{Placement verified.}{%
The evaluator marks the placement as successful.}
\end{samepage}
\smallskip

\begin{samepage}
\appendixtracefinal{The drink has been delivered.}
\end{samepage}

\subsection{Power-Bank Search}
\label{app:cabinet-search-log}

The user asks for a power bank. The power bank is not visible in the scene
graph, and the robot searches the cabinet drawers in order. After finding
the power bank in the lower drawer, the robot delivers it.

\medskip
\noindent{\color{traceRule}\rule{\linewidth}{1.2pt}}\par
\smallskip
\appendixtracerequest{``The computer is out of power. Find me a power bank
and place it on the desk.''}

\begin{samepage}
\appendixtracereasoning{1}{The requested power bank is absent from the visible
scene graph. The drawers in \code{cabinet\_87} may contain objects outside the
current view, so navigate there and inspect them in order.}
\begin{tracecode}
navigate_to(target="cabinet_87")
\end{tracecode}
\appendixtracenavresult{Navigation.}{%
The robot reaches \code{cabinet\_87}.}
\end{samepage}
\smallskip

\begin{samepage}
\appendixtracegroupreasoning{2--7}{The navigation result confirms arrival at
\code{cabinet\_87}. Apply a 0.55\,m approach, a 0.17\,m rightward correction,
and four centimeter-scale adjustments to enter the drawer manipulation pose.}
\appendixtracefold{Base motions}{%
6 moves}
\begin{tracecode}
move_base(direction="forward", distance_m=0.55)
move_base(direction="right", distance_m=0.17)
move_base(direction="forward", distance_m=0.05)
...
\end{tracecode}
\end{samepage}
\smallskip

\begin{samepage}
\appendixtracereasoning{8}{The cabinet is now within manipulation range.
Start with the upper drawer and open it to inspect its contents.}
\begin{tracecode}
open_drawer(drawer_level="higher", container_ref="cabinet_87")
\end{tracecode}
\appendixtraceevaluatorresult{Upper drawer opened.}{%
The evaluator verifies that the upper drawer opens successfully.}
\end{samepage}
\smallskip

\begin{samepage}
\appendixtracereasoning{9}{The refreshed head-camera observation shows the upper
drawer open and the power bank absent. Close the searched compartment before
continuing to avoid leaving an obstruction.}
\begin{tracecode}
close_drawer(
    drawer_level="higher",
    container_ref="cabinet_87")
\end{tracecode}
\appendixtraceevaluatorresult{Upper drawer closed.}{%
The closure is verified successfully.}
\end{samepage}
\smallskip

\begin{samepage}
\appendixtracereasoning{10}{The upper drawer closure is verified. Continue
the ordered search by opening the lower drawer and exposing its contents.}
\begin{tracecode}
open_drawer(drawer_level="lower", container_ref="cabinet_87")
\end{tracecode}
\appendixtraceevaluatorresult{Lower drawer opened.}{%
The evaluator verifies that the lower drawer opens successfully.}
\end{samepage}
\smallskip

\begin{samepage}
\appendixtracereasoning{11}{The refreshed head-camera observation shows the
power bank inside the open lower drawer, and the gripper is empty. The object
is reachable, so pick it and verify control.}
\begin{tracecode}
pick_up(obj="power bank")
\end{tracecode}
\appendixtraceevaluatorresult{Pickup verified.}{%
The wrist observation confirms that the power bank is held securely.}
\end{samepage}
\smallskip

\begin{samepage}
\appendixtracereasoning{12}{The pickup evaluation confirms that the power bank
is held, and the updated scene state removes it from the cabinet. Close the
lower drawer before leaving the search location.}
\begin{tracecode}
close_drawer(drawer_level="lower", container_ref="cabinet_87")
\end{tracecode}
\appendixtraceevaluatorresult{Lower drawer closed.}{%
The cabinet search concludes with both compartments closed.}
\end{samepage}
\smallskip

\begin{samepage}
\appendixtracereasoning{13}{The power bank is held and both drawers are
closed. The destination is \code{desk\_96}; navigate there while retaining the
object.}
\begin{tracecode}
navigate_to(target="desk_96")
\end{tracecode}
\appendixtracenavresult{Navigation.}{%
The robot reaches \code{desk\_96} while holding the power bank.}
\end{samepage}

\begin{samepage}
\appendixtracegroupreasoning{14--16}{The navigation result confirms arrival at
the destination with the power bank held. Apply a 0.80\,m approach followed
by two 0.03\,m corrections to enter the surface-placement envelope.}
\appendixtracefold{Base motions}{%
3 moves}
\begin{tracecode}
move_base(direction="forward", distance_m=0.80)
move_base(direction="forward", distance_m=0.03)
move_base(direction="forward", distance_m=0.03)
\end{tracecode}
\end{samepage}
\smallskip

\begin{samepage}
\appendixtracereasoning{17}{The final base correction has completed, the scene
still marks the power bank as held, and the destination is \code{desk\_96}.
Place the object on its surface and verify the result.}
\begin{tracecode}
place(target="table")
\end{tracecode}
\appendixtraceevaluatorresult{Placement verified.}{%
The power bank is placed successfully on \code{desk\_96}.}
\end{samepage}
\smallskip

\begin{samepage}
\appendixtracefinal{The power bank has been delivered.}
\end{samepage}

\subsection{Desk Cleanup}
\label{app:trash-drop-log}

This episode starts without a user instruction. A standing monitor waits
until the workstation is unoccupied while objects remain on the desk.%
\marginnote{%
  \footnotesize
  \code{monitor\_scene} is an ordinary tool, the standing counterpart of \code{query\_user}: where \code{query\_user} waits for the user to answer, \code{monitor\_scene} waits for the world to change. The agent gives it a natural-language condition, and the call returns once the condition is observed to hold. The returned event then takes the place of a user instruction, giving the task its goal. Only this log uses the tool.%
} \thea loads the workspace-cleaning rules,
identifies an empty orange-juice bottle as trash while leaving the laptop and
charging power bank untouched, and sends one non-blocking notification before
acting. The robot then drops the bottle into the trash bin, and \thea sends a
completion notification.

\medskip
\noindent{\color{traceRule}\rule{\linewidth}{1.2pt}}\par
\smallskip
\noindent{\small\textcolor{traceMuted}{%
\textbf{Standing monitor condition.} ``The workstation is unoccupied, and objects remain
on the desk.''}}\par
\smallskip

\begin{samepage}
\appendixtracereasoning{1}{Keep the desk unchanged while the workstation is
occupied. Set up the standing monitor and wait until the workstation is
unoccupied while items remain on the desk.}
\begin{tracecode}
monitor_scene(
    condition="workstation unoccupied and objects on desk")
\end{tracecode}
\appendixtracenote{traceStageBg}{%
\textbf{\textcolor{traceStageInk}{Monitor event.}}
The user is absent. The desk contains an empty orange-juice bottle, a laptop
connected to a power bank, and several other items.}
\end{samepage}
\smallskip

\begin{samepage}
\appendixtracereasoning{2}{The returned monitor event identifies several desk
objects, but visual presence alone does not justify discarding them. Load the
workspace-cleaning rules.}
\begin{tracecode}
load_skill(name="tidy-workspace")
\end{tracecode}
\appendixtracenote{traceStageBg}{%
\textbf{\textcolor{traceStageInk}{Classification.}}
The confirmed empty bottle is trash. The laptop, charging power bank, and
items whose state is uncertain must remain untouched.}
\end{samepage}
\smallskip

\begin{samepage}
\appendixtracereasoning{3}{The rules and current observation identify only
the empty orange-juice bottle as discardable; the remaining items must stay
untouched. Notify the user, then begin the cleanup.}
\begin{tracecode}
notify_user(
    message="I found an empty orange-juice bottle on your
             desk. I will discard it and leave the other
             items untouched.")
\end{tracecode}
\appendixtracecommresult{Notification.}{%
The one-way message is delivered, and execution continues without awaiting a
reply.}
\end{samepage}

\begin{samepage}
\appendixtracegroupreasoning{4--7}{The target is fixed and the approach is
clear. Advance 0.20\,m, tilt the head by 45 degrees to refresh the local
evidence, then apply 0.05\,m and 0.03\,m corrections to bring the bottle into
the pickup envelope.}
\appendixtracefold{Base motions}{%
4 actions}
\begin{tracecode}
move_base(direction="forward", distance_m=0.20)
tilt_head(pitch_deg=45)
move_base(direction="forward", distance_m=0.05)
...
\end{tracecode}
\end{samepage}
\smallskip

\begin{samepage}
\appendixtracereasoning{8}{The refreshed view and incremental alignment place
the confirmed empty bottle within reach while the gripper remains empty.
Execute the grasp and verify control.}
\begin{tracecode}
pick_up(obj="empty orange-juice bottle")
\end{tracecode}
\appendixtraceevaluatorresult{Pickup verified.}{%
The wrist observations show the bottle secured between the gripper fingers.}
\end{samepage}
\smallskip

\begin{samepage}
\appendixtracereasoning{9}{The grasp evaluator shows the empty bottle secured
between the gripper fingers. The disposal destination is
\code{trash\_can\_54}; navigate there while retaining the bottle.}
\begin{tracecode}
navigate_to(target="trash_can_54")
\end{tracecode}
\appendixtracenavresult{Navigation.}{%
The robot reaches \code{trash\_can\_54} with the bottle.}
\end{samepage}
\smallskip

\begin{samepage}
\appendixtracegroupreasoning{10--12}{The navigation result confirms arrival in
the target area, but the bin is not yet centered for release. Apply a
0.55\,m approach, a 15-degree left turn, and a final 0.05\,m correction to
reach the release pose.}
\appendixtracefold{Base motions}{%
3 moves}
\begin{tracecode}
move_base(direction="forward", distance_m=0.55)
move_base(direction="turn_left", angle_deg=15)
move_base(direction="forward", distance_m=0.05)
\end{tracecode}
\end{samepage}

\begin{samepage}
\appendixtracereasoning{13}{The empty bottle remains held,
\code{trash\_can\_54} is unambiguous, and the final alignment succeeds.
Release the bottle into the bin and verify the resulting state.}
\begin{tracecode}
trash_drop(trash_ref="trash_can_54")
\end{tracecode}
\appendixtraceevaluatorresult{Drop verified.}{%
The bin opening is centered, both grippers are empty, and no bottle is visible
outside the bin.}
\end{samepage}
\smallskip

\begin{samepage}
\appendixtracereasoning{14}{The drop evaluator confirms that both grippers are
empty and no bottle is visible outside the bin, satisfying the cleanup goal.
Notify the user that the bottle has been discarded.}
\begin{tracecode}
notify_user(
    message="The empty orange-juice bottle has been
             discarded.")
\end{tracecode}
\appendixtracecommresult{Completion notification.}{%
The cleanup result is delivered to the user.}
\end{samepage}
\smallskip

\begin{samepage}
\appendixtracefinal{The empty orange-juice bottle has been discarded.}
\end{samepage}

\subsection{Basket Delivery}
\label{app:g1-basket-delivery-log}

The user (Jason) asks for snacks to be brought to friends. The robot packs snacks
into a basket, carries it into the living room, walks up to Jason, and hands
over the basket.%
\marginnote{%
  \footnotesize
  Routine low-level calls (controller checks, camera refreshes, setup
  queries, and the locomotion steps inside navigation) are folded and
  excluded from the turn numbering.%
}

\medskip
\noindent{\color{traceRule}\rule{\linewidth}{1.2pt}}\par
\smallskip
\appendixtracerequest{``I am about to meet some friends. Bring some snacks
over for them.''}

\begin{samepage}
\appendixtracereasoning{1}{The workspace is ready. Place the snack
in the basket and wait for the manipulation to finish before lifting the
basket.}
\begin{tracecode}
pack_snacks()
\end{tracecode}
\appendixtracefoldin{traceManipBg}{traceManipInk}{Packing workflow}{repeated controller updates folded}
\appendixtraceevaluatorresult{Placement verified.}{%
The snack is packed in the basket.}
\end{samepage}
\smallskip

\begin{samepage}
\appendixtracereasoning{2}{Packing is complete. Lift the loaded
basket, retract the arm for walking, and prepare it for navigation.}
\begin{tracecode}
lift_basket()
\end{tracecode}
\appendixtraceevaluatorresult{Lift verified.}{%
The basket is ready to be carried.}
\end{samepage}

\begin{samepage}
\appendixtracereasoning{3}{The basket is secured for transport.
Follow the doorway route into the living room before selecting a recipient.}
\begin{tracecode}
navigate_to(living_room)
\end{tracecode}
\appendixtracenavresult{Navigation.}{%
G1 reaches the living room while holding the basket.}
\end{samepage}
\smallskip

\begin{samepage}
\appendixtracereasoning{4}{The living-room view shows Jason seated in a
white shirt and light trousers. Navigate to him while retaining the basket.}
\begin{tracecode}
navigate_to(Jason)
\end{tracecode}
\appendixtracenavresult{Navigation.}{%
G1 brings Jason into front view while retaining the basket.}
\end{samepage}

\begin{samepage}
\appendixtracegroupreasoning{5--8}{The recipient is now in front but not yet
close enough for the handoff. Continue in short increments.}
\appendixtracefold{Carry motions}{4 moves}
\begin{tracecode}
basket_carry_move(direction="forward", distance_m=3.0)
basket_carry_move(direction="forward", distance_m=1.5)
...
\end{tracecode}
\appendixtracenavresult{Navigation.}{%
Four short forward motions bring G1 close to the seated recipient.}
\end{samepage}
\smallskip

\begin{samepage}
\appendixtracereasoning{9}{The recipient is close in front of G1, and the
basket remains held. Extend the arm to the handoff pose and open the hand, but
do not infer transfer from the controller state alone.}
\begin{tracecode}
hand_over()
\end{tracecode}
\appendixtracemanipresult{Handoff.}{%
The right arm extends, and the hand opens successfully.}
\appendixtraceevaluatorresult{Release verified.}{%
The basket is released and rests securely with the recipient.}
\end{samepage}

\endgroup

\end{document}